\documentclass[notitlepage, 12pt]{article}

\usepackage[utf8]{inputenc}
\usepackage{xcolor}
\usepackage{amsmath}
\usepackage{microtype}
\usepackage{balance}
\usepackage{amssymb}
\usepackage{amsthm}
\usepackage{graphicx}
\usepackage[font=footnotesize,labelfont=bf]{caption}
\usepackage{anyfontsize}
\usepackage{wrapfig}
\usepackage{titlesec}
\usepackage[font=footnotesize]{subcaption}

\newcommand{\eps}{\varepsilon}

\newcommand{\HH}{\mathcal{H}}
\newcommand{\Fi}{\mathbb{F}}

\newcommand\newsubcap[1]{\phantomcaption%
\caption*{\textbf{\figurename~\thefigure\thesubfigure:} #1}}

\titleformat{\section}
  {\normalfont\scshape}{\thesection}{1em}{}
\titleformat{\subsection}
  {\normalfont\scshape}{\thesubsection}{1em}{}

\title{Detecting unusual input to neural networks}
\author{Jörg Martin, Clemens Elster \\ Physikalisch-Technische Bundesanstalt (PTB)}
\begin{document}

\twocolumn[
\begin{@twocolumnfalse}
	\maketitle
	\begin{abstract}
		Evaluating a neural network on an input that differs markedly from the training data might cause erratic and flawed predictions. We study a method that judges the unusualness of an input by evaluating its informative content compared to the learned parameters. 
		This technique can be used to judge whether a network is suitable for processing a certain input and to raise a red flag that unexpected behavior might lie ahead.
		We compare our approach to various methods for uncertainty evaluation from the literature for various datasets and scenarios. Specifically, we introduce a simple, effective method that allows to directly compare the output of such metrics for single input points even if these metrics live on different scales.
	\end{abstract}
\end{@twocolumnfalse}
]

\section{Introduction}
\label{sec:introduction}
\begin{figure*}[t]\centering
	\begin{subfigure}[t]{0.32\textwidth}
		\centering
		\includegraphics[width=\textwidth]{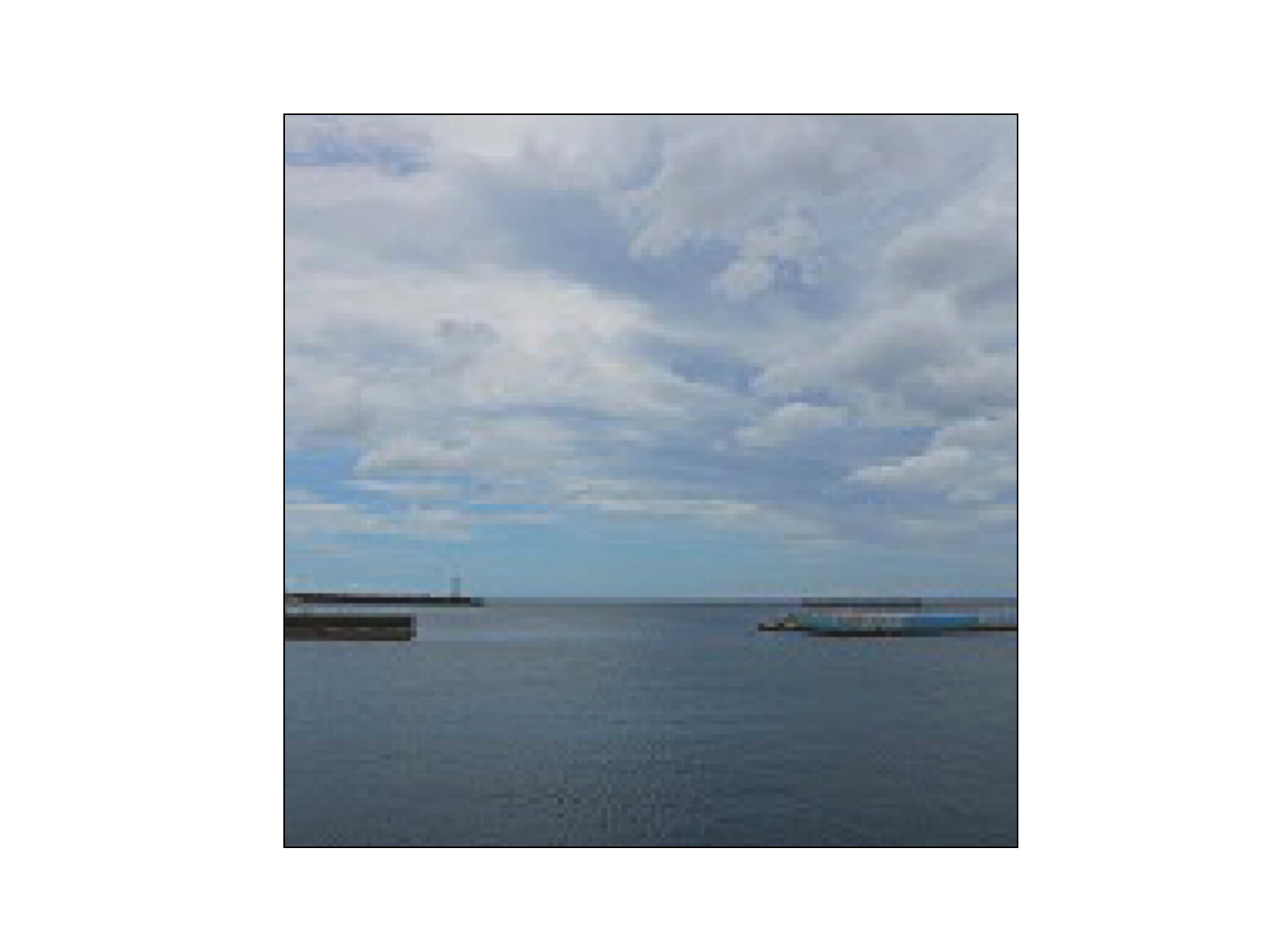}
		\caption*{
			\textit{prediction}: ``sea'' \\
			$1-p(\hat{y}|x): \,\,\,\,\,1.2\,\%$ \\
		$\langle\mathcal{F}_\theta(x)\rangle: \qquad 35.0\, \%$}
		\label{subfig:mountain}
	\end{subfigure}%
	~
	\begin{subfigure}[t]{0.32\textwidth}
		\centering
		\includegraphics[width=\textwidth]{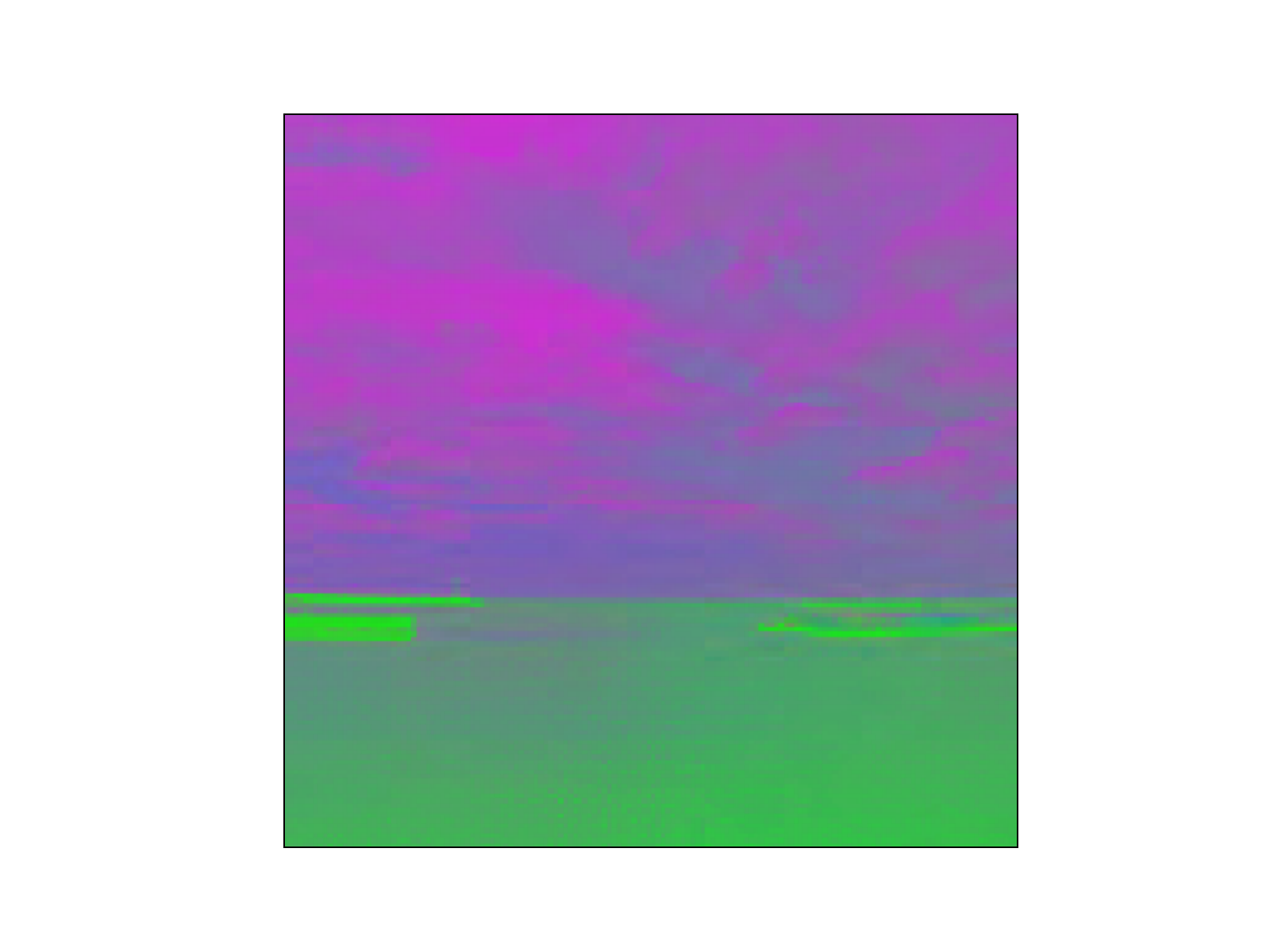}
		\caption*{``sea''\\
		$14.1\%$\\
		$98.2\%$}
		\label{eq:full_monochrome_mountain}
	\end{subfigure}%
	~
	\begin{subfigure}[t]{0.32\textwidth}
		\centering
		\includegraphics[width=\textwidth]{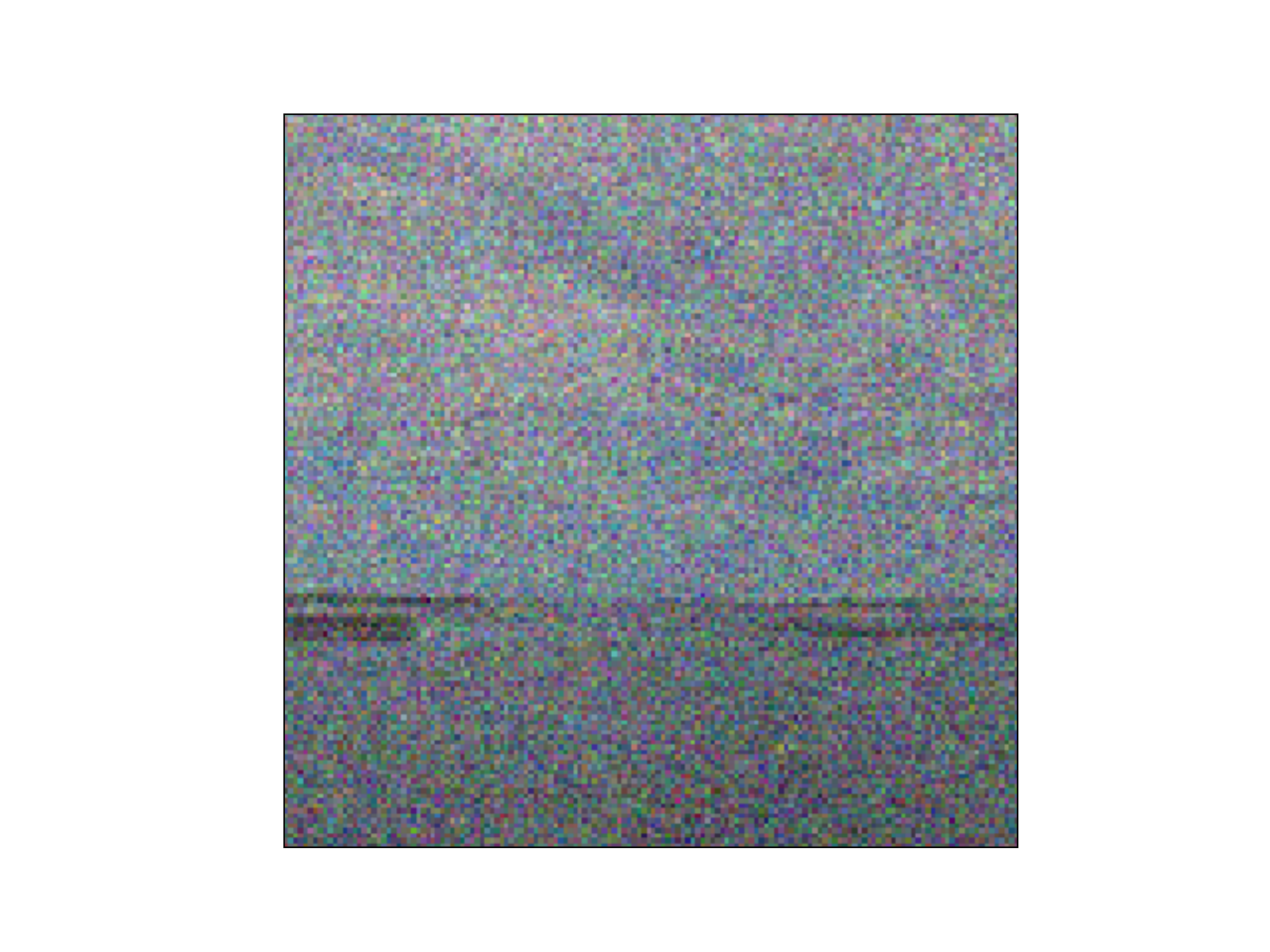}
		\caption*{``forest''\\
		$2.1\%$\\
		$97.2\%$}
		\label{subfig:monochrome_mountain}
	\end{subfigure}
	\caption{An image showing the sea from the Intel Image Classification dataset \cite{IntelImageClassification} in its original form (left), after inverting its green color channel in RGB format (middle) and after adding Gaussian noise (right). Below we denote the prediction $\hat{y}$ of the neural network from Section \ref{subsec:modified_data} together with its \emph{error probability} $1-p(\hat{y}|x)$ and the quantity $\langle\mathcal{F}_\theta(x)\rangle$ based on the \emph{Fisher information} we introduce in Section \ref{sec:method}. We observe in this example that the naive error probability can be misleading, while the Fisher information indicates an unusual input due to its value close to 100\%.}
	\label{fig:intro_monochrome}
\end{figure*}The reliability and performance of a machine learning algorithm depends crucially on the data that were used for training it \cite{Cortes1995}. Having an incomplete training set or encountering input data that have unprecedented deviations might lead to unexpected, erroneous behavior \cite{Amodei2016, GooglePhotos, Marcus2018}. In this work we consider the task of judging whether an input to a neural network, trained for classification, is \emph{unusual} in the sense that it is different from the training data. 
For this purpose we consider a quantity based on the Fisher information matrix \cite{Cover2012}. A similar quantity was shown by the authors to be useful for the related task of detecting adversarial examples \cite{Martin2020}. Two kinds of scenarios are studied:
\begin{itemize}
	\item The input data were \emph{modified}, e.g. by being noisy or contorted.
	\item The training set is \emph{incomplete}, by missing some structural component that will emerge in practice.
\end{itemize}
Our considerations are carried out along various datasets and in comparison with related methods from the literature. In particular, we present an easy and efficient method to directly compare different metrics that judge the uncertainty behind a prediction.

We consider neural networks that are trained for classifying inputs into $C$ classes $1,\ldots,C$. After application of a softmax activation the output of a neural network for an input $x$ can be read as a vector of probabilities $(p_\theta(y|x))_{y=1,\ldots,C}$, where we wrote $\theta$ for the parameters of the neural network (usually weights and biases) \cite{Bishop2006}. The class $\hat{y}$ where this vector is maximal is then the prediction of the network. It is quite tempting to consider $p_\theta(\hat{y}|x)$ as a measure for the confidence behind this prediction.  The leftmost column of Figure \ref{fig:intro_monochrome} shows an image from the Intel Image Classification dataset \cite{IntelImageClassification} and the output of a neural network that was trained on it, details will follow in Section \ref{sec:experiments} below. The image is classified correctly as ``sea'' with $p(\hat{y}|x)=98.8\%$. In this article we are more interested in the probability of \emph{misclassification}
\begin{align}
\label{eq:intro_error_prob}
 1- p(\hat{y} | x) = 1.2 \% \,.
\end{align}
One might expect that the higher this probability, the more unreliable is the prediction of the network. The second column of Figure \ref{fig:intro_monochrome} shows the same image but with with an inverted green color channel. While the classification remains``sea'', the probability of error rises to $14.1\%$ which appears to be a sensible behavior.  Unfortunately, the behavior of the output probability is not always meaningful.  The rightmost column of Figure \ref{fig:intro_monochrome} shows the same picture with Gaussian noise added. 
The image is misclassified as showing a ``forest'' but with a quite small error probability $1-p(\hat{y} | x )$ of around $2.1\%$.
The fact that the softmax probability is not fully exhaustive or even misleading in judging the reliability of the prediction of the network led to various developments in the literature like a Bayesian judgement of uncertainty \cite{Gal2016, Kingma2015, Kingma2013, Gal2016Thesis, Gal2017, Blundell2015, Filos2019} and  deep ensembles \cite{Lakshminarayanan2017}. We here study the behavior of a quantity, called the \emph{Fisher form} and denoted by $\mathcal{F}_\theta$, that evaluates if the input $x$ differs from what the network ``has learned''. $\mathcal{F}_\theta(x)$ was introduced by the authors, with slight modifications, for the purpose of adversarial detection \cite{Szegedy2013, Goodfellow2014} for the first time in \cite{Martin2020}.
Below Figure \ref{fig:intro_monochrome} we listed the values of $\langle\mathcal{F}_\theta(x) \rangle$, where $\langle\ldots\rangle$ denotes some normalization we introduce below. As we see in Figure \ref{fig:intro_monochrome} $\langle\mathcal{F}_\theta(x)\rangle$ shows a quite natural behavior on the depicted images, where values close to 100\% indicate an unusual input.

This article is structured as follows: In Section \ref{sec:method} we introduce, and motivate, the method used in this article and introduce the normalization that allows us to compare different metrics. Section \ref{sec:experiments} presents the results for various datsets and splits in two halfs: in Subsection \ref{subsec:modified_data} we study the effect of modifying the input, while Subsection \ref{subsec:incomplete_training_data} considers the case where the training data lack some structural aspects. Finally we provide some conclusions and an outlook for future research.

\section{Method}

\label{sec:method}
\begin{figure*}[t]\centering
\begin{subfigure}[t]{0.32\textwidth}
	\centering
	\includegraphics[width=\textwidth]{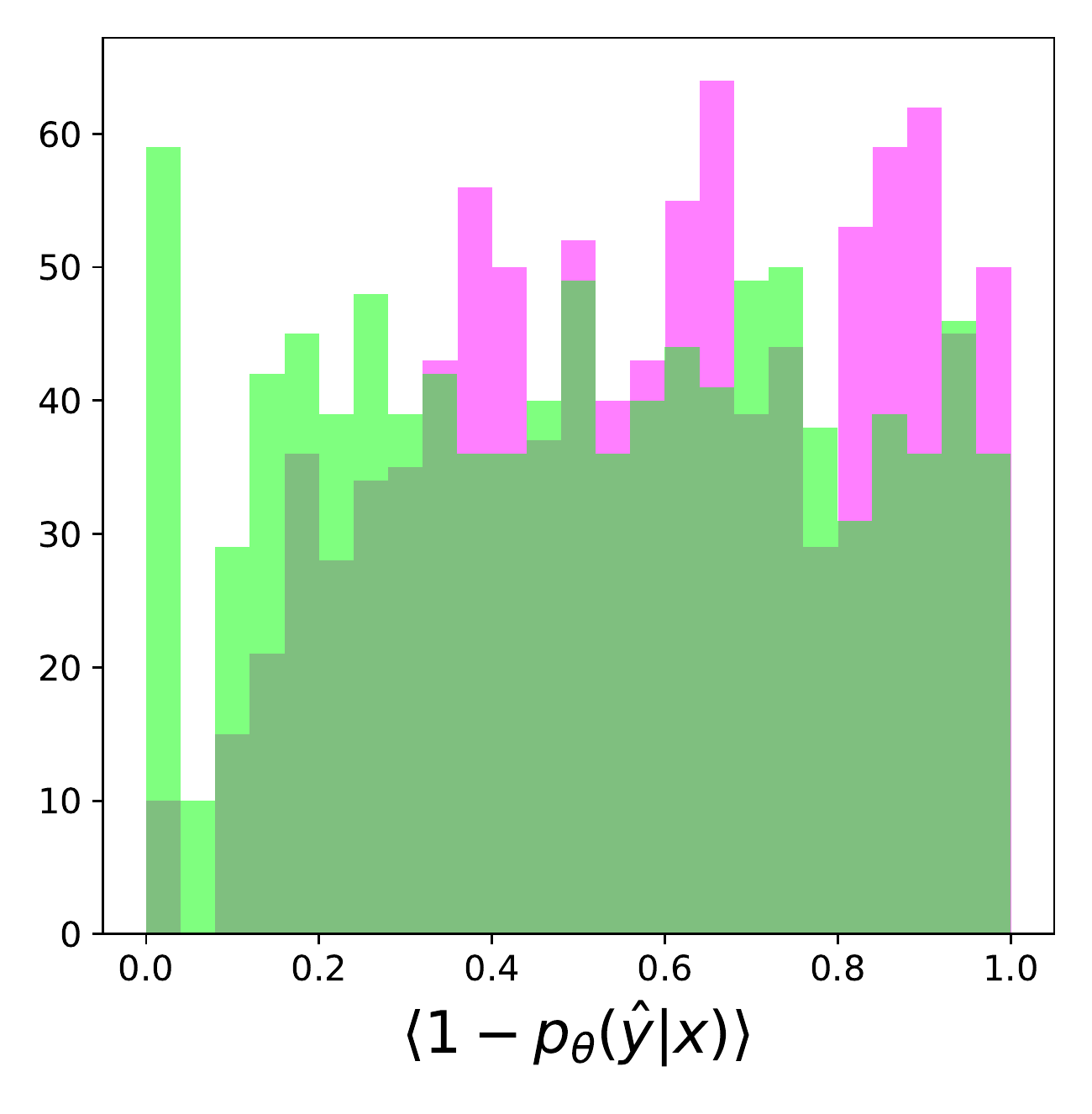}
	\end{subfigure}%
	~
	\begin{subfigure}[t]{0.32\textwidth}
		\centering
		\includegraphics[width=\textwidth]{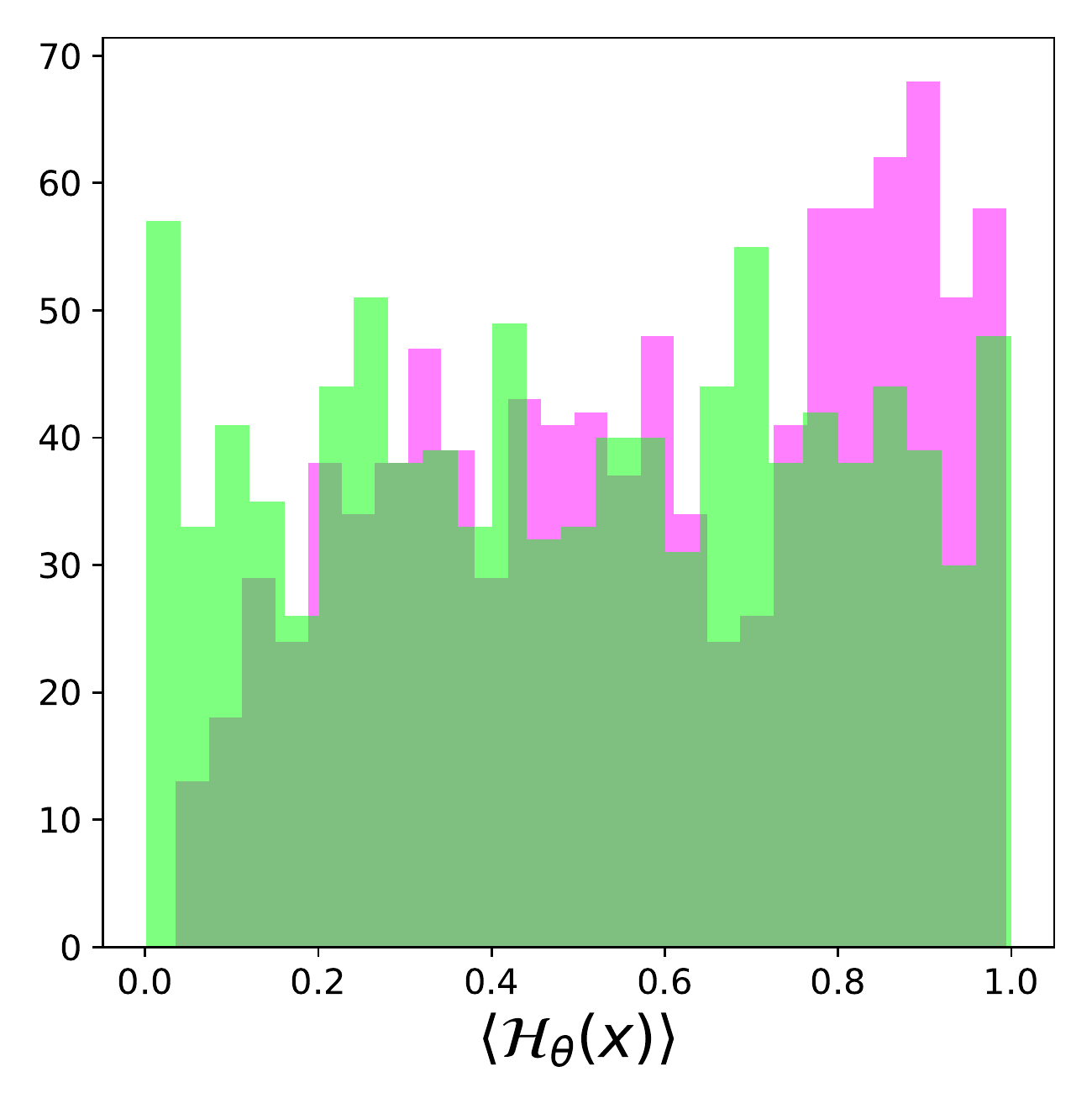}
	\end{subfigure}%
	~
	\begin{subfigure}[t]{0.32\textwidth}
		\centering
		\includegraphics[width=\textwidth]{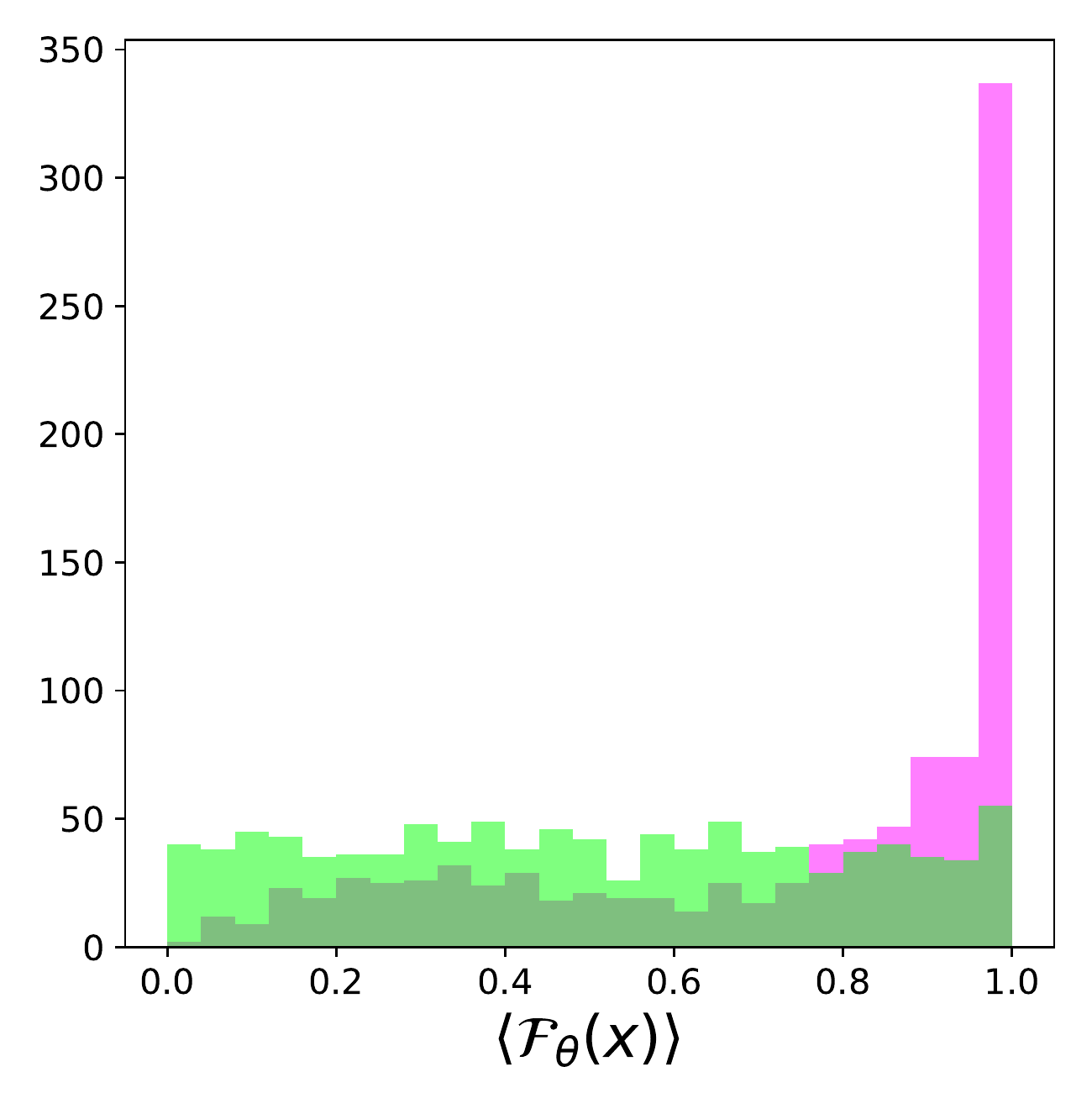}
	\end{subfigure}
	\caption{Using the normalization \eqref{eq:normalization} we can directly compare the behavior of $1-p(\hat{y}|x)$ from \eqref{eq:intro_error_prob}, the entropy $\mathcal{H}_\theta(x)$ from \eqref{eq:entropy} and the Fisher form $\mathcal{F}_\theta(x)$ from \eqref{eq:FisherForm} and ignore the fact that these quantities live on different scales. The histograms depict the behavior of these quantities for images from the Intel Image Classification dataset \cite{IntelImageClassification} unmodified (in green) and after inverting the green color channel (in magenta) as in the middle of Fig. \ref{fig:intro_monochrome}.}  
	\label{fig:intel_monochrome_hist}
\end{figure*}
A slightly more complete summary of the output probabilities than \eqref{eq:intro_error_prob} can be gained by considering the \emph{entropy} instead \cite{Cover2012, Gal2016Thesis}:
\begin{align}
	\label{eq:entropy}
	\mathcal{H}_\theta(x) {=} {-}\!\sum_{y=1}^C p_\theta(y|x) \log p_\theta(y|x).
\end{align}
If the entropy is high the output probabilities $(p(y|x))_{y=1,\ldots, C}$ are roughly equal in magnitude and the prediction is uncertain. 
$\mathcal{H}_\theta(x)$ suffers still from the same drawback than $p_\theta(\hat{y} | x)$: Specifically, it completely ignores the uncertainties of $\theta$ and uses their point estimates only. One way around this problem is to use another, more elaborate distribution such as an approximate to the posterior predictive or a mixed distribution from a deep ensemble.

In \cite{Martin2020} the authors introduced a quantity that gives a deeper insight than \eqref{eq:entropy} and is based on the Fisher information. The Fisher information matrix for a neural network with output $p_\theta(y|x)$ for an input $x$ and a $p$-dimensional parameter $\theta$ equals $\Fi_\theta(x) = \sum_{y=1}^C \nabla_\theta p_\theta(y|x) \cdot \nabla_\theta \log p_\theta(y|x)^T\,. $
Even a smaller network has around $p=10^6$ parameters, which makes $\Fi_\theta$ a matrix with $10^{12}$ entries and is thus, in practice, infeasible. A way out, proposed in \cite{Martin2020}, is to consider the effect of this matrix in a specific direction $v$, i.e. to consider the quadratic form 
\begin{align}
	\label{eq:FisherForm}
	\mathcal{F}_\theta(x) = v^T \Fi_\theta(x) v  \,,
\end{align}
where $v$ has the same dimensionality as $\theta$. We will refer to \eqref{eq:FisherForm} as the \emph{Fisher form}. The quantity \eqref{eq:FisherForm} measures \emph{how much information is gained/lost} in a (small) step in the direction $v$. More precisely, the Kullback-Leibler divergence between $p_{\theta}(y|x)$ and  $p_{\theta +\eps v}(y|x)$ can be written as $\frac{\eps^2}{2}\, \mathcal{F}_\theta(x)$$+$$\mathcal{O}(\eps^3)$ \cite{Kullback1997}. After the parameter $\theta$ is learned, this divergence will be large for $x$ that are informative, that is \emph{different from those used to infer the learned value of $\theta$}. To use \eqref{eq:FisherForm} we have to fix a suitable direction $v$. A natural choice, which we will use throughout this article, is the negative gradient of the entropy
\begin{align}
\label{eq:v}
v = -  \partial_\theta \HH_\theta(x) \,,
\end{align}
where $\HH_\theta(x)$ is as in \eqref{eq:entropy}. The motivation behind a choice like \eqref{eq:v} is as follows: Once trained the network will produce for an input the highest probability for one class $\hat{y}$, its ``prediction'', and lower probabilities for all other classes $y \neq \hat{y}$. The vector $v$ then denotes the direction in parameter space that decreases the entropy, in other words that tends to increase the probability for $\hat{y}$ further and to decrease all other probabilities.

If the input is unusual and leads to a wrong classification $\hat{y}$, then a step in the direction of $v$ will increase the information substantially, as the used $x$ will usually be completely different from the training data that were used for inferring $\theta$. For another choice of $v$ compare \cite{Martin2020}.

Taking the expression $\mathcal{F}_\theta(x)$ has the useful consequence that we can rewrite the quadratic form in \eqref{eq:FisherForm} as 
$
	\mathcal{F}_\theta(x) {=} \sum_{y=1}^C \! D_v p_\theta(y|x)\, D_v\! \log p_\theta(y|x)
$,
where $D_v f(\theta)=v^T \nabla_\theta f(\theta)$ denotes the directional derivative w.r.t $\theta$ in direction of $v$, which can be computed either directly or using a finite difference approximation that avoids the need for backpropagation \cite{Martin2020}.

Before analyzing how the quantity \eqref{eq:FisherForm} performs in practice, there is one final point that needs some consideration. While quantities such as the Fisher form \eqref{eq:FisherForm} or the entropy \eqref{eq:entropy} can each be compared for two different datapoints there is no direct possibility for a comparison between the value of the entropy and the Fisher information for the same datapoint. 
Moreover, requiring someone applying these quantities to get an intuition first before they can judge whether a quantity is ``high enough'' to rise suspicion is rather unsatisfactory. 

When faced with a binary decision problem, such as ``will produce unexpected behavior or not'', we can use the \emph{receiver operator characteristic} (ROC) and the associated \emph{area under the ROC curve} (AUC) to compare different metrics, cf., for example \cite{Martin2020}. However, a comparison based on the ROC/AUC still has the drawback that it is not applicable for \emph{single datapoints} and, moreover, that not all problems can be seen as a binary decision. We here describe a rather easy, yet efficient approach that solves these issues. Let us introduce the following \emph{normalization}
for a quantity $q(x)$ and a test set $T$ 
\begin{align}
	\label{eq:normalization}
	\langle q(x) \rangle = \frac{\#\{x'\mbox{ in $T$ s.t. } q(x') < q(x)\}}{\#\{x'\mbox{ in $T$}\}},
\end{align}
where $\#$ counts the number of elements. In other words $\langle q(x) \rangle$ denotes fraction of test samples $x'$ in $T$ for which $q(x') < q(x)$. By construction, $\langle q(x) \rangle$ depends on the input $x$ and on the test set $T$ that is used. This normalization has a few nice properties: It is 
\begin{itemize}
	\item \emph{bounded}, as it lies always between 0 and 1,
	\item \emph{monotone}: from $q(x_1) < q(x_2)$ it follows $\langle q(x_1) \rangle < \langle q(x_2) \rangle$,
	\item \emph{invariant} under strictly monotone transformations such as taking a logarithm or scaling.
\end{itemize}
\begin{figure*}[t]
\begin{subfigure}[t]{0.49\textwidth}\centering
	\begin{minipage}[t]{\textwidth}
		\centering
		\includegraphics[width=\textwidth]{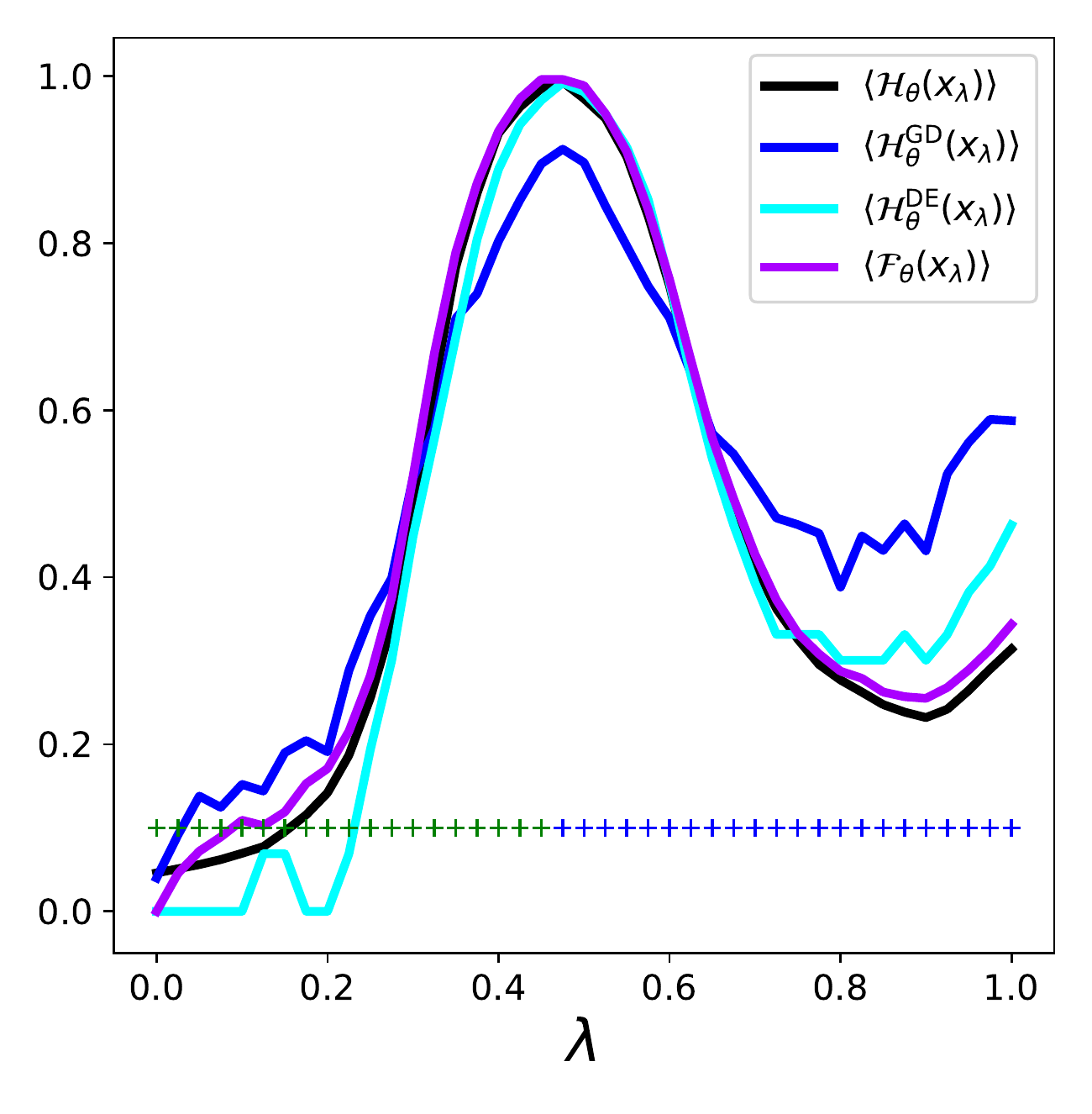}
	\end{minipage}\\
	\begin{minipage}[t]{0.19\textwidth}
		\centering
		\includegraphics[width=\textwidth]{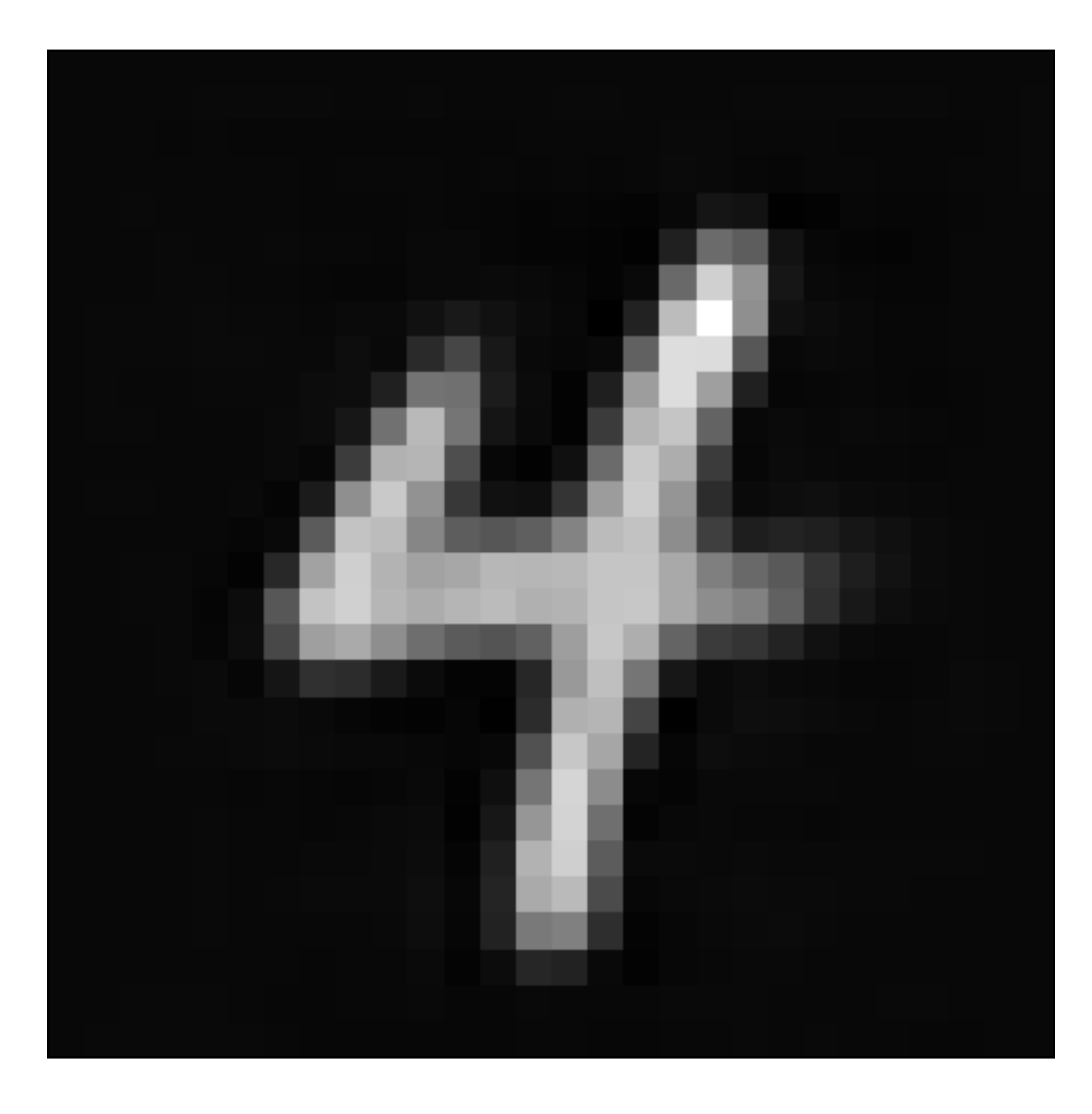}
		\caption*{$\lambda=0$}
	\end{minipage}%
	~
	\begin{minipage}[t]{0.19\textwidth}
		\centering
		\includegraphics[width=\textwidth]{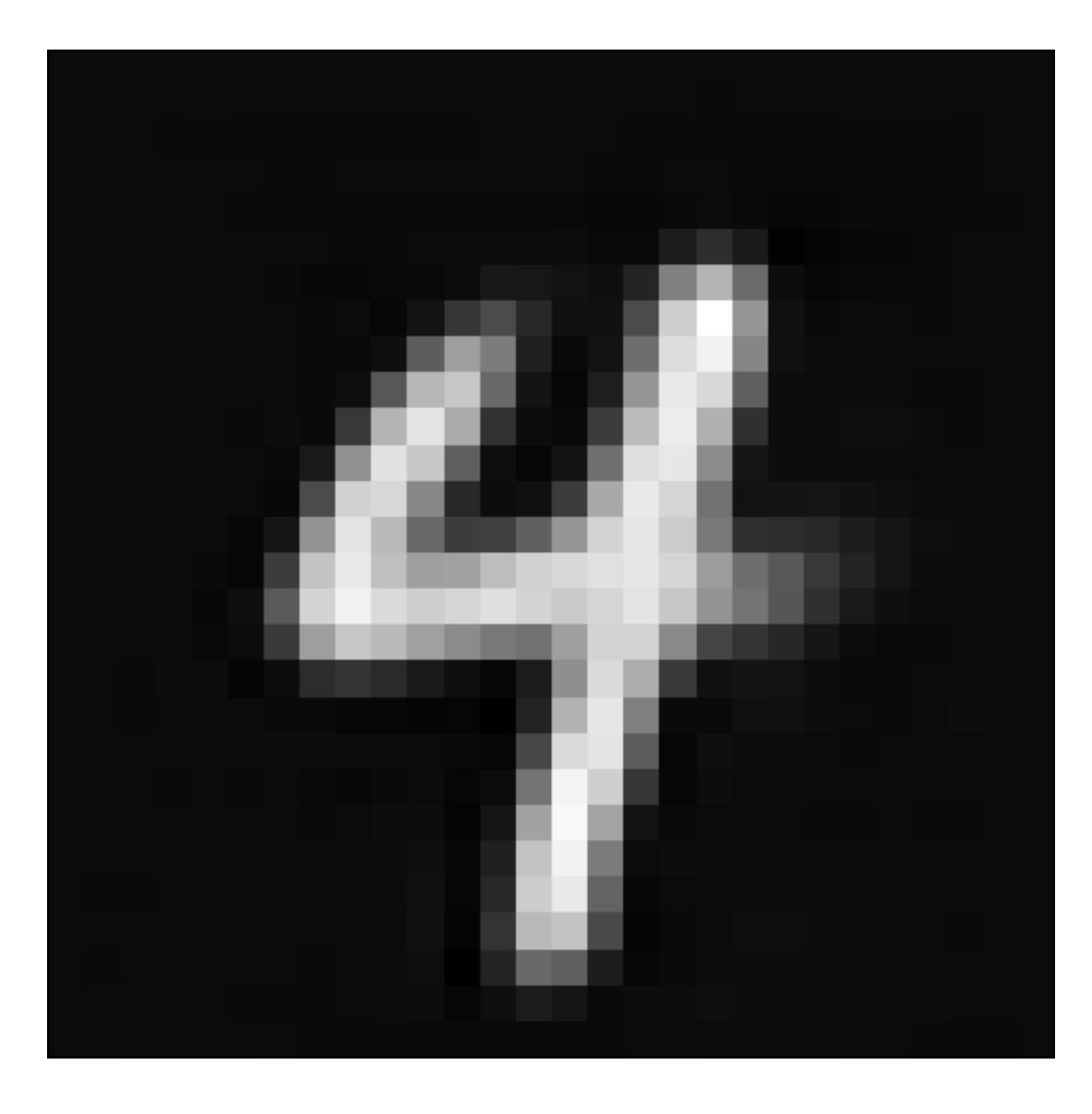}
		\caption*{$0.25$}
	\end{minipage}%
	~
	\begin{minipage}[t]{0.19\textwidth}
		\centering
		\includegraphics[width=\textwidth]{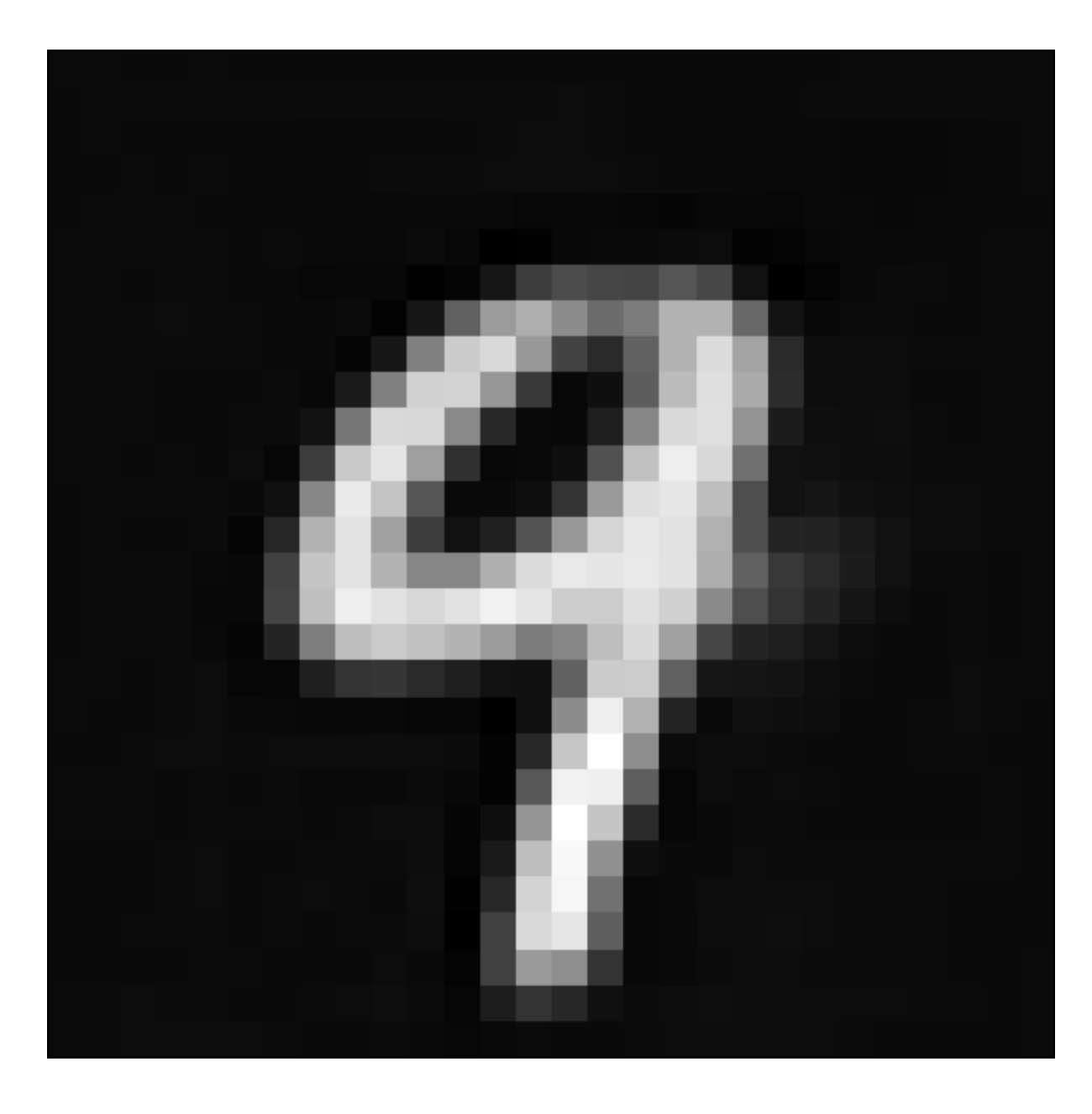}
		\caption*{$0.50$}
	\end{minipage}%
	\begin{minipage}[t]{0.19\textwidth}
		\centering
		\includegraphics[width=\textwidth]{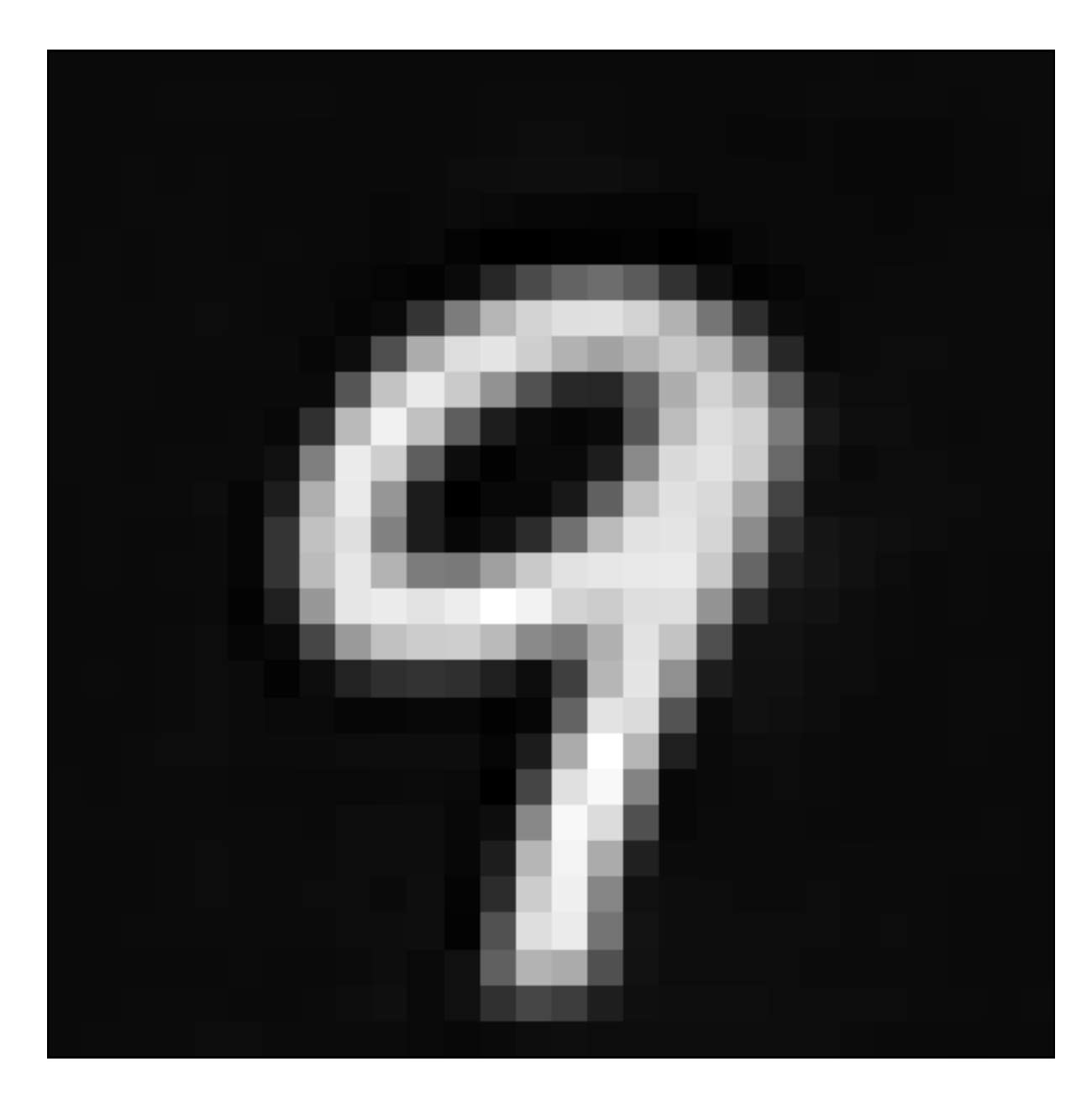}
		\caption*{$0.75$}
	\end{minipage}%
	\begin{minipage}[t]{0.19\textwidth}
		\centering
		\includegraphics[width=\textwidth]{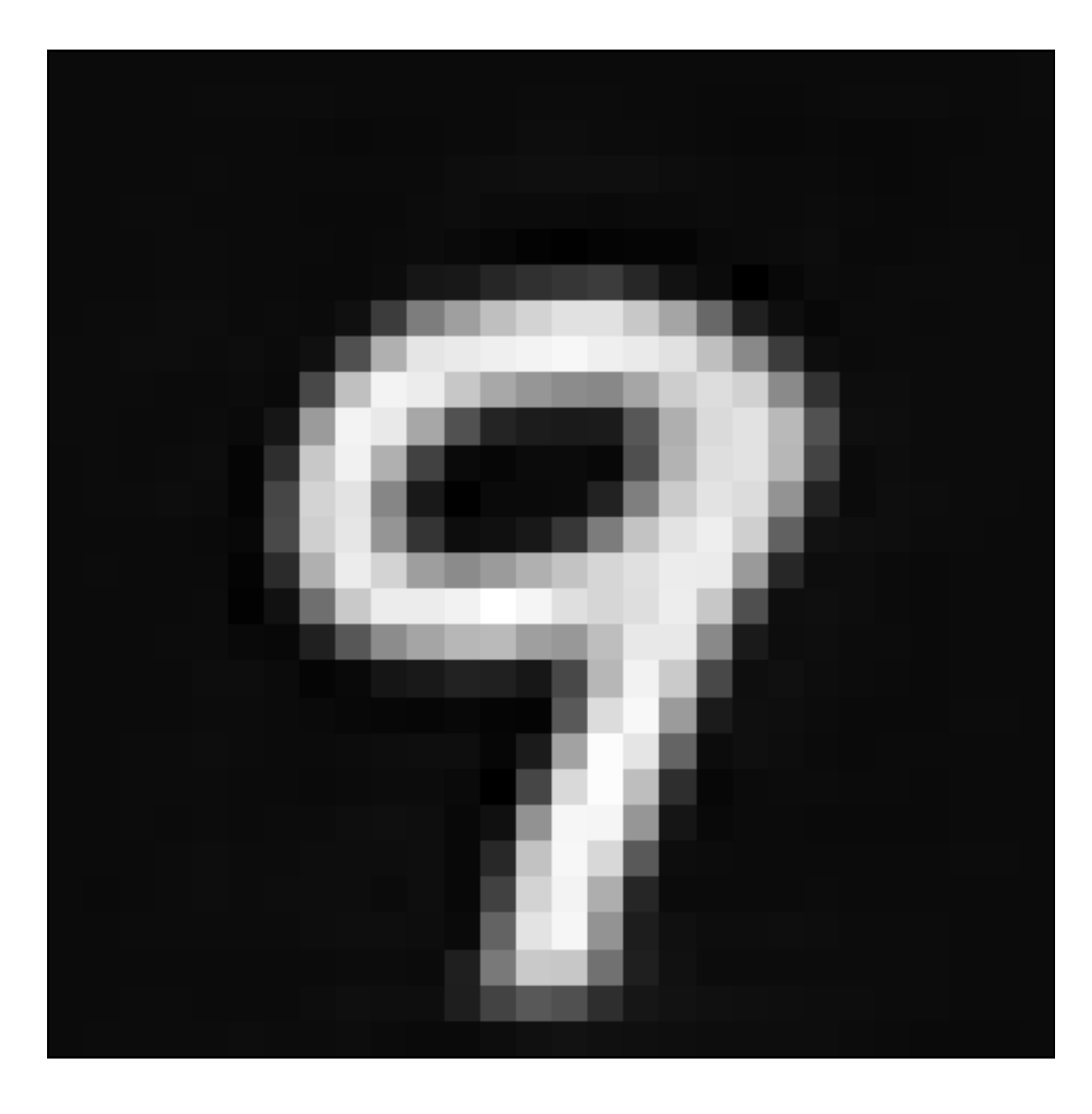}
		\caption*{$1$}
	\end{minipage}
	\newsubcap{Behavior of the normalized \emph{entropies} $\langle\mathcal{H}_\theta(x_\lambda)\rangle$, $\langle\mathcal{H}^{\mathrm{GD}}_\theta(x_\lambda)\rangle$, $\langle\mathcal{H}^{\mathrm{DE}}_\theta(x_\lambda)\rangle$ for the softmax output, Gaussian dropout and a deep ensemble (in black, blue and cyan) and the \emph{Fisher form} $\langle \mathcal{F}_\theta(x_\lambda) \rangle$ (in purple) when interpolating two images from the MNIST dataset \cite{LeCun2010} using a variational autoencoder. $x_\lambda$ and $\lambda$ are as in \eqref{eq:x_lambda_mnist}. The small crosses indicate the classification as 4 (in green) and 9 (in blue)}
	\label{fig:mnist_interpolation}
\end{subfigure}
\begin{subfigure}[t]{0.49\textwidth}\centering
	\begin{minipage}[t]{\textwidth}
		\centering
		\includegraphics[width=\textwidth]{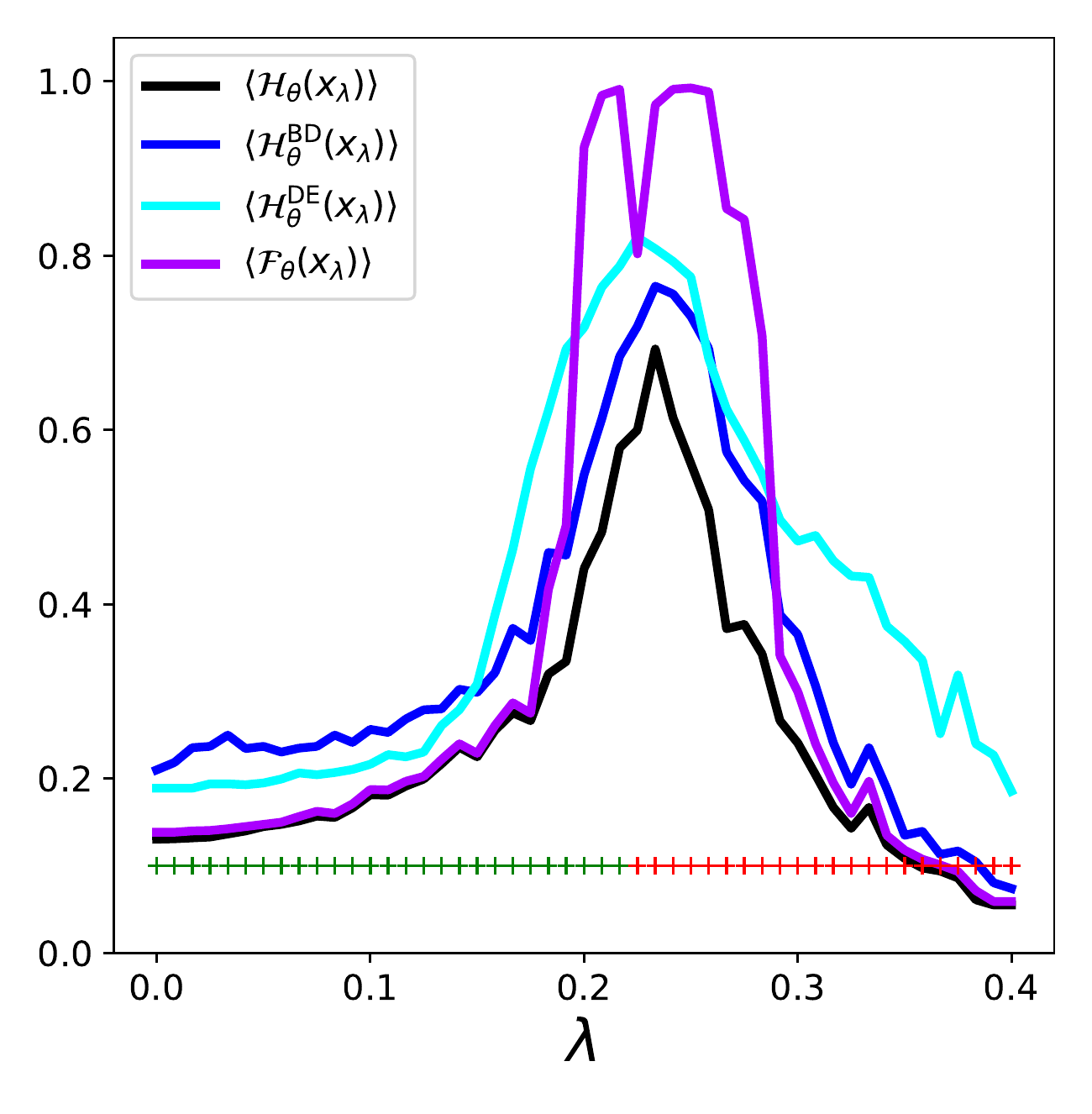}
	\end{minipage}\\
	\begin{minipage}[t]{0.19\textwidth}
		\centering
		\includegraphics[width=\textwidth]{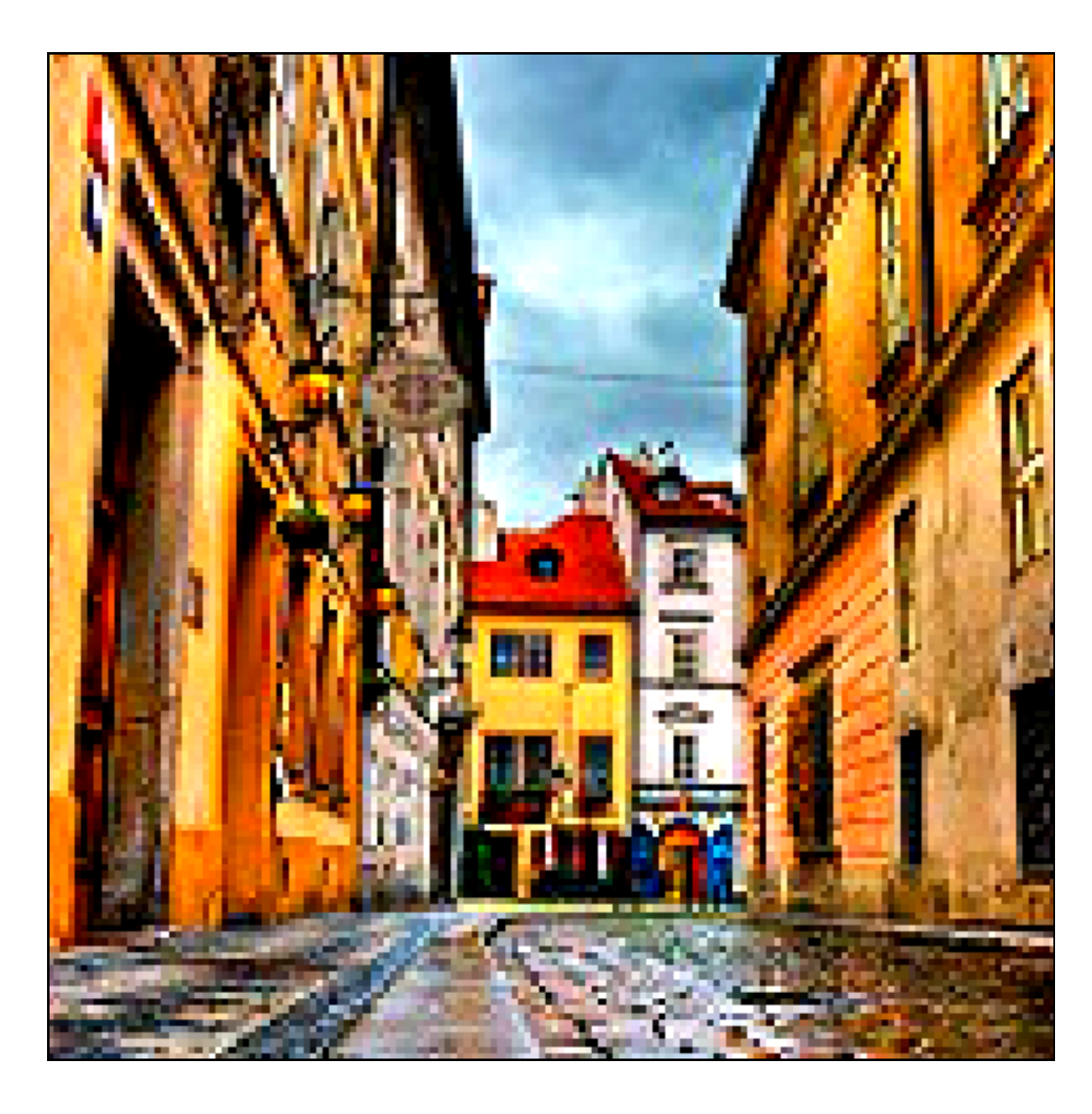}
		\caption*{$\lambda=0$}
	\end{minipage}%
	~
	\begin{minipage}[t]{0.19\textwidth}
		\centering
		\includegraphics[width=\textwidth]{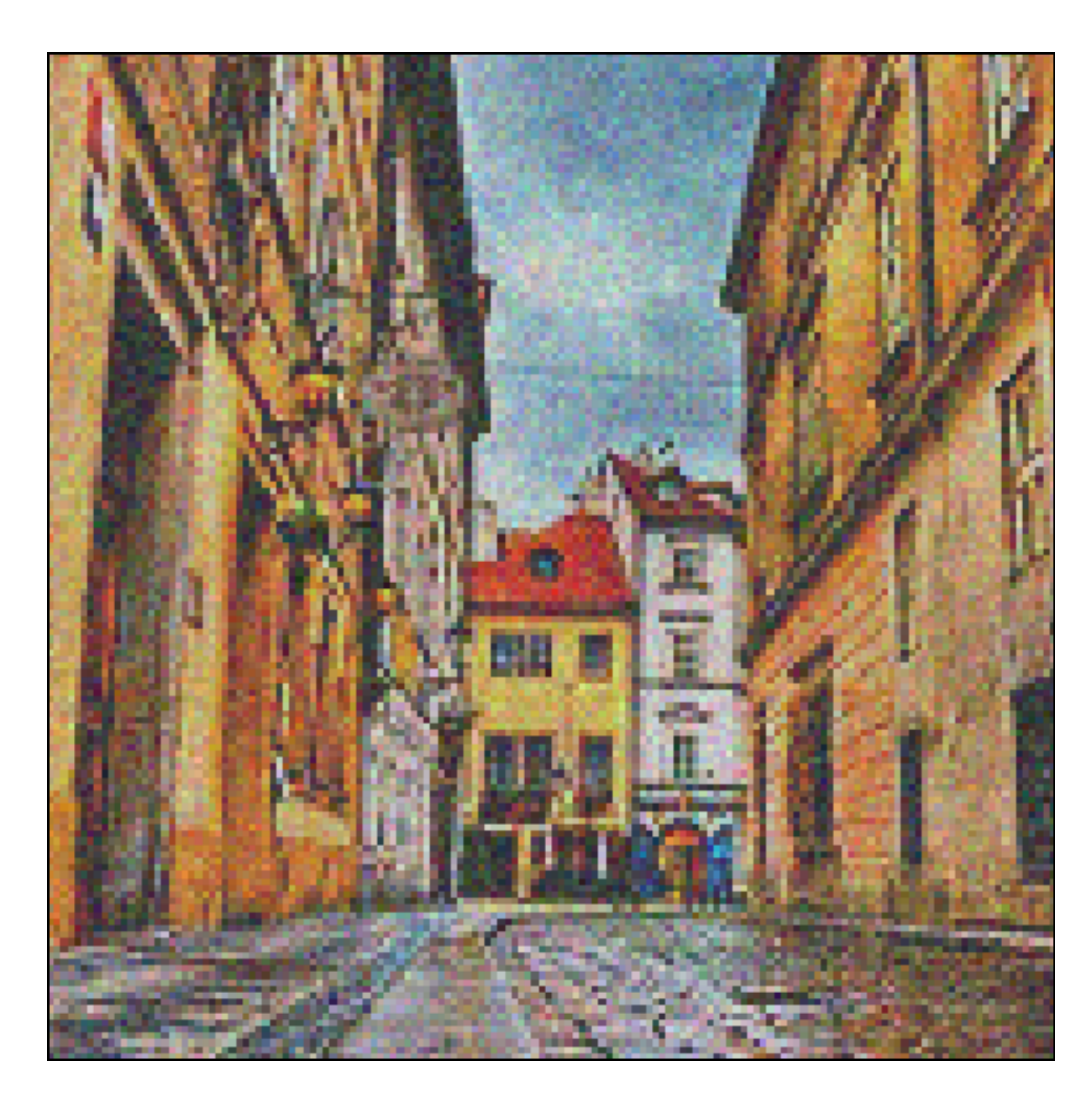}
		\caption*{$0.10$}
	\end{minipage}%
	~
	\begin{minipage}[t]{0.19\textwidth}
		\centering
		\includegraphics[width=\textwidth]{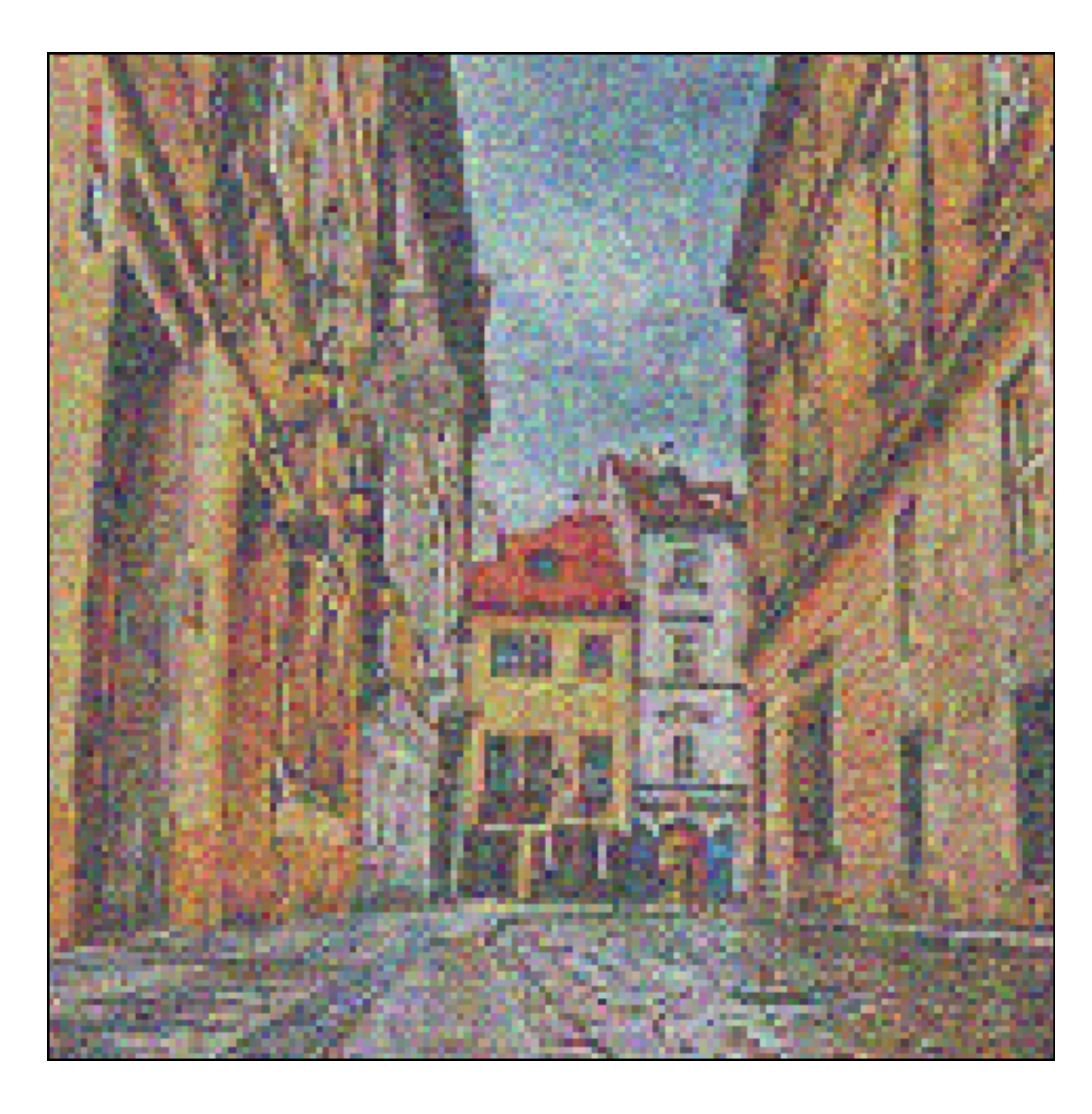}
		\caption*{$0.20$}
	\end{minipage}%
	\begin{minipage}[t]{0.19\textwidth}
		\centering
		\includegraphics[width=\textwidth]{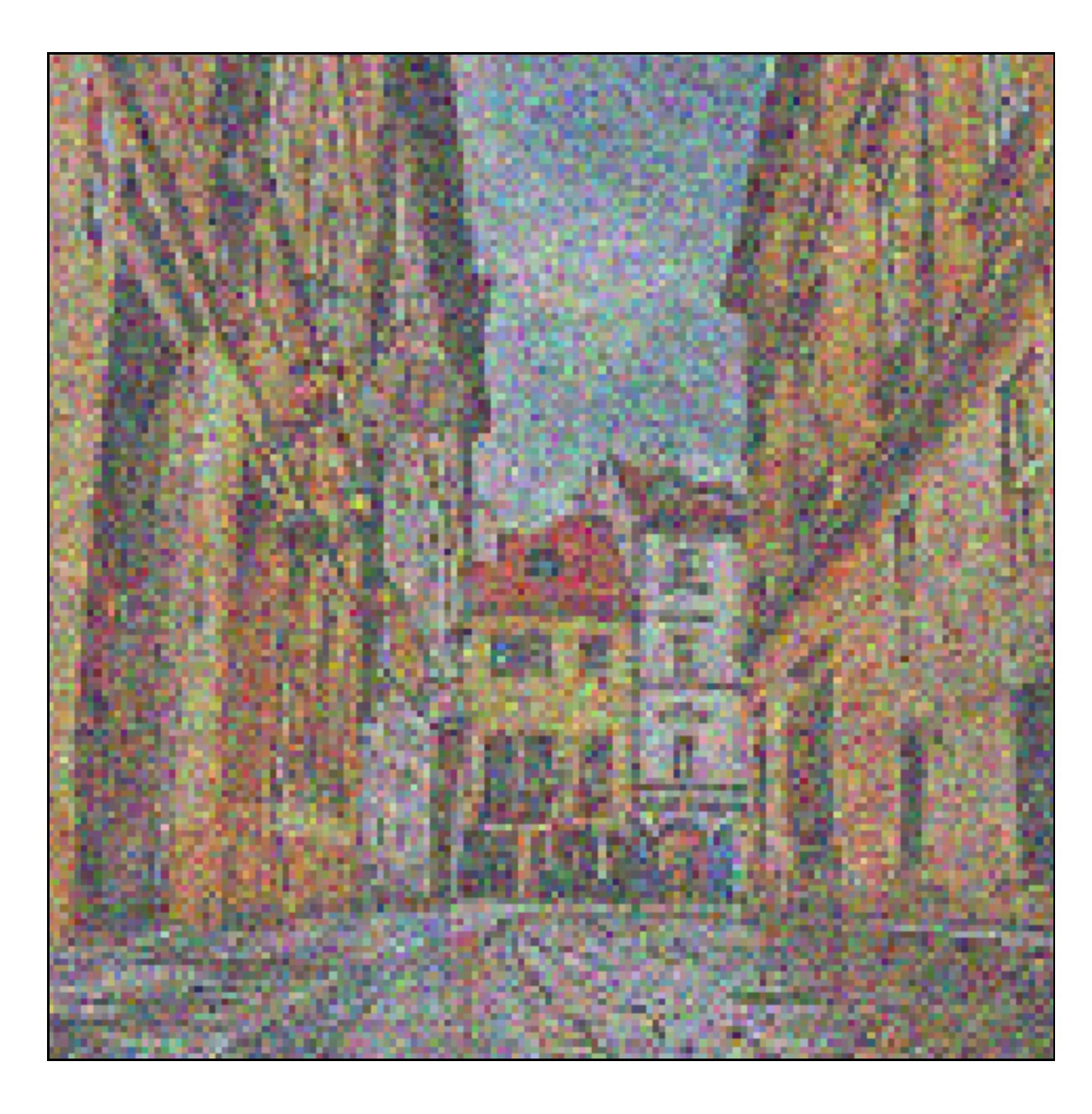}
		\caption*{$0.30$}
	\end{minipage}%
	\begin{minipage}[t]{0.19\textwidth}
		\centering
		\includegraphics[width=\textwidth]{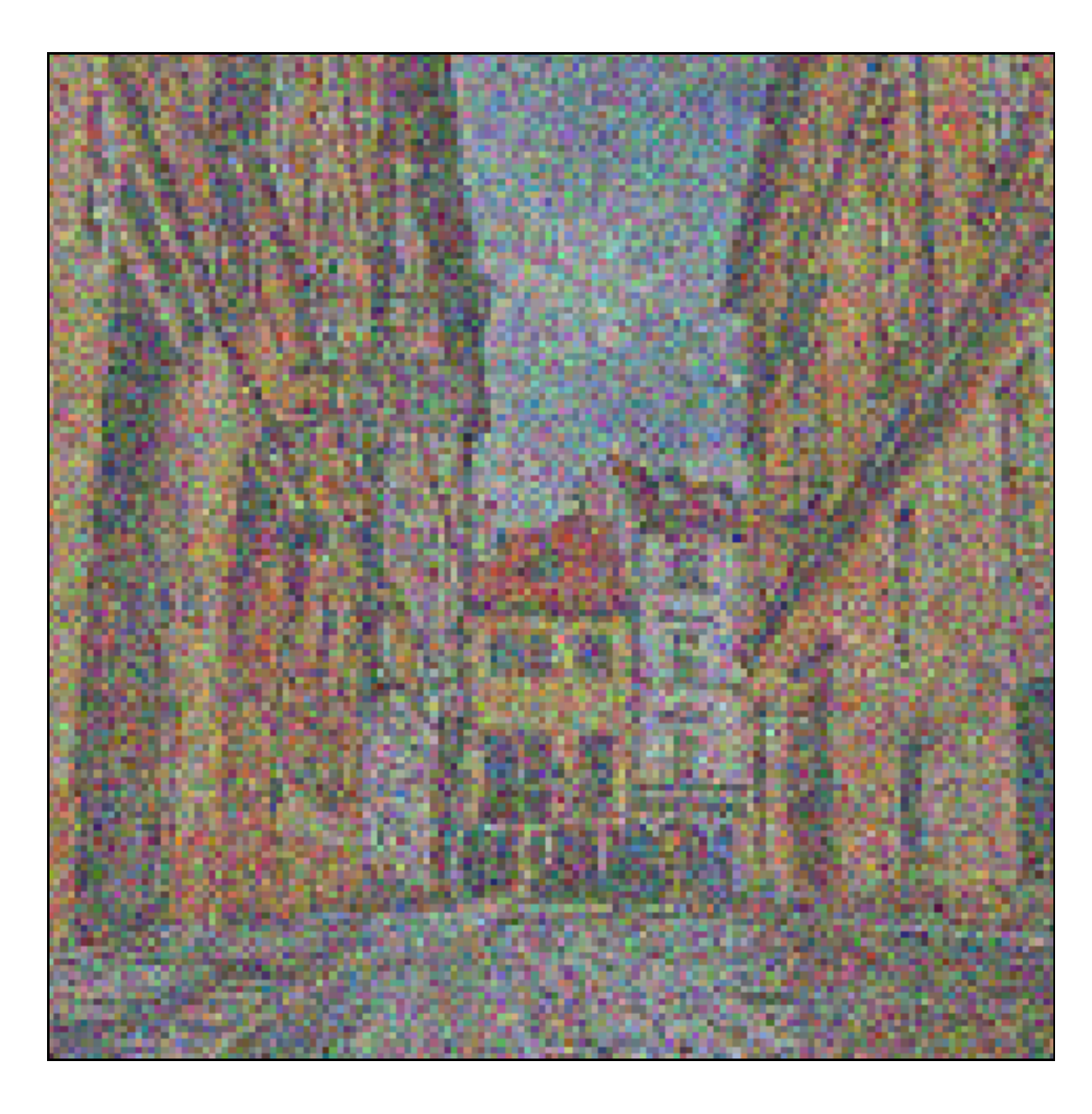}
		\caption*{$1$}
	\end{minipage}
	\newsubcap{Behavior of the normalized \emph{entropies} $\langle\mathcal{H}_\theta(x_\lambda)\rangle$, $\langle\mathcal{H}^{\mathrm{GD}}_\theta(x_\lambda)\rangle$, $\langle\mathcal{H}^{\mathrm{DE}}_\theta(x_\lambda)\rangle$ for the softmax output, Bernoulli dropout and a deep ensemble (in black, blue and cyan) and the \emph{Fisher form} $\langle \mathcal{F}_\theta(x_\lambda) \rangle$ (in purple) when pertubing an image from the Intel Image Classification dataset by Gaussian noise. $x_\lambda$ and $\lambda$ are as in \eqref{eq:x_lambda_intel}. The crosses indicate whether the street has been classified correctly (in green) or not (in red).}
	\label{fig:intel_noise}
\end{subfigure}
\end{figure*}
This normalization allows us to compare different measures of reliability/ uncertainty, since relations such as 
$ \langle \mathcal{F}_\theta(x) \rangle =75\% \mbox{ and } \langle \HH_\theta(x)\rangle=75\% $
have a similar meaning: $75\%$ of the test samples have a smaller value than the one for $x$. If we see the magnitude of a quantity such as $\mathcal{F}_\theta(x)$ as an indicator for the ``unusualness'' of a datapoint $x$, then a value of $\langle \mathcal{F}_\theta(x)\rangle$ close to 100\% can be seen as a strong indication that the reliability for a prediction based on $x$ will be quite questionable. In particular a relation such as 
$ \langle \mathcal{F}_\theta(x) \rangle > \langle \mathcal{H}_\theta(x) \rangle $
for an $x$ that causes a wrong prediction will indicate that $\mathcal{F}_\theta(x)$ detected the underlying uncertainty more clearly than $\mathcal{H}_\theta(x)$ (based on the test set), so that a normalization via $\langle\ldots \rangle$ make measures of reliability directly comparable.

Figure \ref{fig:intel_monochrome_hist} shows the distribution of $\langle 1-p_\theta(\hat{y}|x) \rangle$ (left), the normalized version of \eqref{eq:intro_error_prob}, the normalized entropy $\langle \mathcal{H}_\theta(x) \rangle$ (middle) and Fisher form $\langle\mathcal{F}_\theta(x) \rangle$ (right) for the same neural network as in Figure \ref{fig:intro_monochrome} for the original images (in green) and those with inverted green color channel (in magenta). 
The test set for the normalization was chosen equal to the test set from \cite{IntelImageClassification}. Note that only the Fisher information has a distinct peak at high values for the modified images.

\section{Experiments}
\label{sec:experiments}
In this section we discuss the behavior of the Fisher form $\mathcal{F}_\theta$, the entropy $\mathcal{H}_\theta$, the entropy $\mathcal{H}_\theta^{\mathrm{DE}}$ for the mixture distribution predicted by a deep ensemble \cite{Lakshminarayanan2017} and the entropy $\mathcal{H}_\theta^{\mathrm{GD}/\mathrm{BD}}$ of an approximate to the  posterior predictive.
More precisely, $\mathcal{H}_\theta^{\mathrm{DE}}$ refers to the average of entropies produced by each member of the ensemble, and $\mathcal{H}_\theta^{\mathrm{GD}/\mathrm{BD}}$ to the mean of entropies obtained by repeated predictions of the trained net using dropout. 
All deep ensembles were constructed using 5 networks, trained independently of each other. To approximate the posterior predictive we used Gaussian dropout \cite{Kingma2015} for MNIST and Bernoulli dropout \cite{Gal2016} (with rate 0.5) for all other examples, as the latter seemed to converge substantially faster. The entropy for the distribution produced by the Gaussian dropout will be denoted by $\mathcal{H}_\theta^{\mathrm{GD}}$ and we will write $\mathcal{H}_\theta^{\mathrm{BD}}$ for Bernoulli dropout. The architecture of all used networks is sketched in the Appendix of this article.

In Section \ref{subsec:modified_data} we will analyze the behavior of $\mathcal{F}_\theta, \mathcal{H}_\theta, \mathcal{H}_\theta^{\mathrm{DE}}$ and $\mathcal{H}_\theta^{\mathrm{GD}/\mathrm{BD}}$ on datapoints that are modified, either by adding noise or by a transformation based on a variational autoencoder. Section \ref{subsec:incomplete_training_data} studies the effect of missing certain features while training.

\subsection{Modified data}
\label{subsec:modified_data}

As a first example we will consider the MNIST dataset \cite{LeCun2010} for digit recognition. We trained a convolutional neural network (Fig. \ref{fig:mnist_net}) for 25 epochs until it reached an accuracy of around $99\%$ on both, the test and training set. 
Finally we trained a \emph{variational autoencoder} (VAE) \cite{Kingma2013}, sketched in Figure \ref{fig:mnist_vae}, for 100 epochs until it reached a mean squared error accuracy of around 1\% of the pixel range. Using a VAE is a popular form for interpolating between datapoints \cite{Berthelot2018, Smith2018}. 
To this end, we split the VAE into an encoder $E$ and a decoder $D$. For single input images $\tilde{x}_0$ and $\tilde{x}_1$ we then take $ z_0 = E(\tilde{x}_0)$, $z_1 = E(\tilde{x}_1 ) $
and set for $\lambda$ between 0 and 1:
\begin{align}
	\label{eq:x_lambda_mnist}
	x_\lambda = D(z_0 + \lambda (z_1 - z_0))\,.
\end{align}
Figure \ref{fig:mnist_interpolation} shows the interpolation between an image showing a 4 and an image showing a 9 for different values of $\lambda$. The curves above the reconstructed images show the dependency of $\langle \mathcal{H}_\theta(x_\lambda) \rangle$ (in black), $\langle \mathcal{F}_\theta(x_\lambda) \rangle$ (in magenta), the Gaussian dropout entropy $\langle \mathcal{H}_\theta^{\mathrm{GD}}(x_\lambda) \rangle$ (in blue) and the deep ensemble entropy $\langle \mathcal{H}_\theta^{\mathrm{DE}}(x_\lambda) \rangle$ (in cyan) on $\lambda$.
We used the MNIST test set as the set $T$ for normalization. All methods show a similar behavior with a distinct peak in the transition phase.

\begin{figure}[h]
	\includegraphics[width=0.48\textwidth]{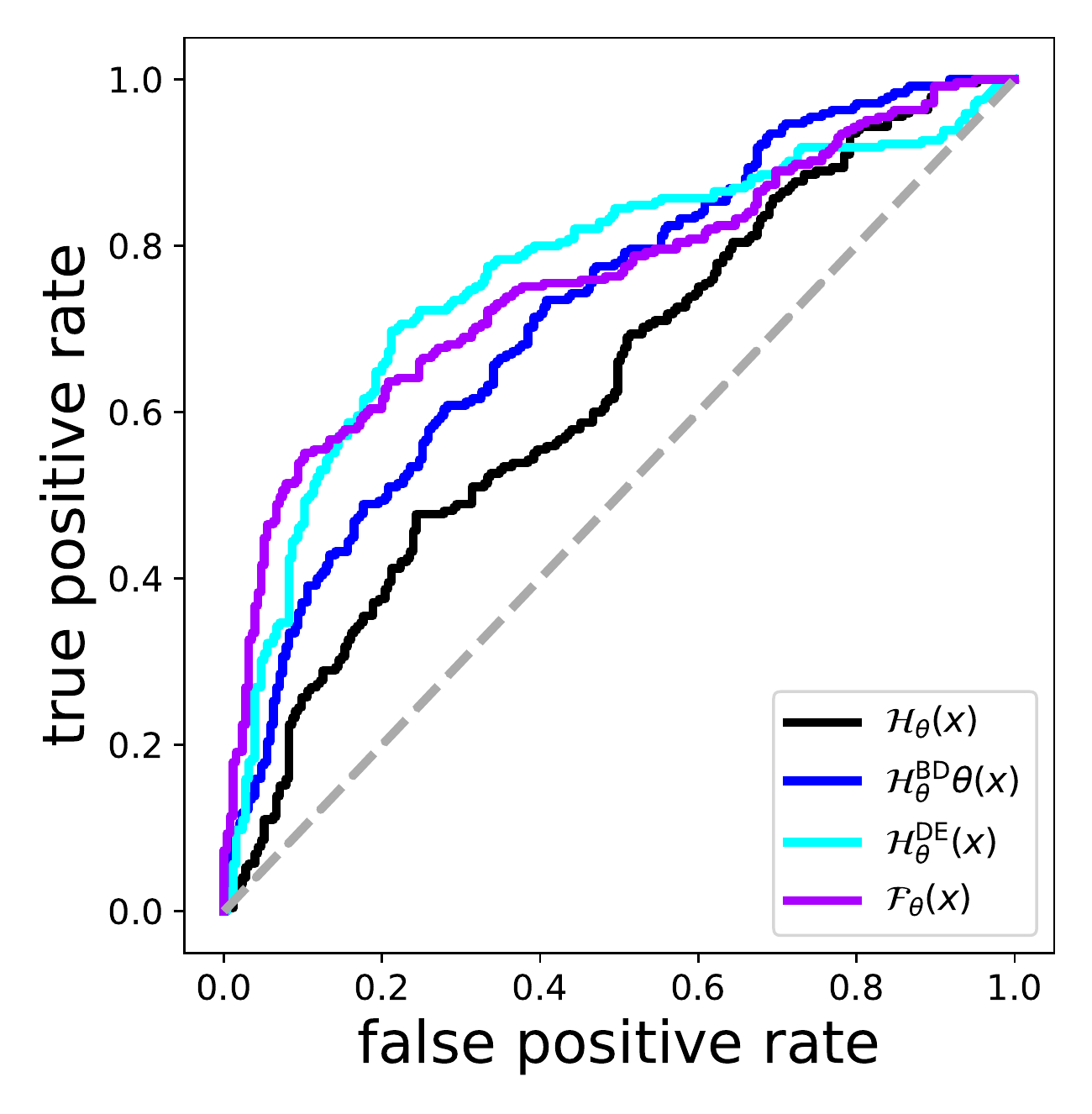}
	\caption{The ROC for detecting whether an image from the Intel Image Classification dataset was modified by inverting its green color channel (as in the middle of Figure \ref{fig:intro_monochrome}) via the entropies $\mathcal{H}_\theta(x)$, $\mathcal{H}_\theta^{\mathrm{BD}}(x)$, $\mathcal{H}_\theta^{\mathrm{DE}}(x)$ (in black, blue and cyan) and the Fisher form $\mathcal{F}_\theta(x)$ (in purple)}
	\label{fig:intel_roc}
\end{figure}

As a next dataset we will consider the \emph{Intel Image Classification} dataset \cite{IntelImageClassification} that contains images of 6 different classes (street, building, forest, mountain, glacier, sea). We touched this example already in the introduction of this article. We trained a convolutional neural network (Fig. \ref{fig:credit_net}) on this dataset for 15 epochs until it reached an accuracy of around $80\%$ on the test set of \cite{IntelImageClassification}, which we also used for normalization.
We here want to consider the effect of two modifications on the datapoints. The effect of inverting the green color channel was already shown in Figure \ref{fig:intro_monochrome} and Figure \ref{fig:intel_monochrome_hist}. This modification lets the accuracy of the network drop from around 80\% to 30\%. Figure \ref{fig:intel_roc} shows the ROC for $ \mathcal{H}_\theta $ $\mathcal{F}_\theta$, $\mathcal{H}_\theta^{\mathrm{DE}}$ and $\mathcal{H}_\theta^{\mathrm{BD}}$ (now with Bernoulli dropout) for detecting such modified images, where we used once more the test set as reference. Note that a normalization as in \eqref{eq:normalization} is not needed as the ROC is invariant under monotone transformations. The AUC for $\mathcal{H}_\theta$ and $\mathcal{H}_\theta^{\mathrm{BD}}$ are 0.63 and 0.72, while $\mathcal{H}_\theta^{\mathrm{DE}}$ and $\mathcal{F}_\theta$ both share an AUC of $0.76$.

Next, we consider the effect of noise. The upper plot in Figure \ref{fig:intel_noise} shows the evolution of $\langle \mathcal{H}_\theta(x_\lambda) \rangle $, $\langle \mathcal{H}_\theta^{\mathrm{BD}}(x_\lambda) \rangle $, $\langle \mathcal{H}_\theta^{\mathrm{DE}}(x_\lambda) \rangle $ and $\langle \mathcal{F}_\theta (x_\lambda) \rangle$ where $x_\lambda$ is now given as 
\begin{align}
	\label{eq:x_lambda_intel}
	x_\lambda = x + \lambda \cdot \eps\,,
\end{align}
with $x$ being a single image from the Intel Image Classification dataset that shows a street and with $\eps$ being an array of standard normal noise of the same shape as $x$. The lower row of Figure \ref{fig:intel_noise} displays $x_\lambda$ for various $\lambda$. 

The crosses in the upper plot in Figure \ref{fig:intel_noise} depict if the network classifies the image correctly as a street (in green) or whether it chooses the wrong class (in red). We observe that while all quantities $\langle \mathcal{H}_\theta(x_\lambda)\rangle$, $\langle \mathcal{F}_\theta (x_\lambda) \rangle$, $\langle\mathcal{H}_\theta^{\mathrm{DE}}(x_\lambda)\rangle$ and $\langle\mathcal{H}_\theta^{\mathrm{BD}}(x_\lambda)\rangle$ share a similar trend, the Fisher form, is well above all other quantities in the transition phase where the noise starts flipping the classification. Inspecting the maximal Softmax output $p(\hat{y}|x_\lambda)$ shows that beyond $\lambda =0.3$ this probability is almost identical to 1, which finally pushes all quantities down towards 0.

\subsection{Incomplete training data}
\label{subsec:incomplete_training_data}
\begin{figure*}[t]\centering
	\begin{subfigure}[t]{0.45\textwidth}
		\centering
		\includegraphics[width=\textwidth]{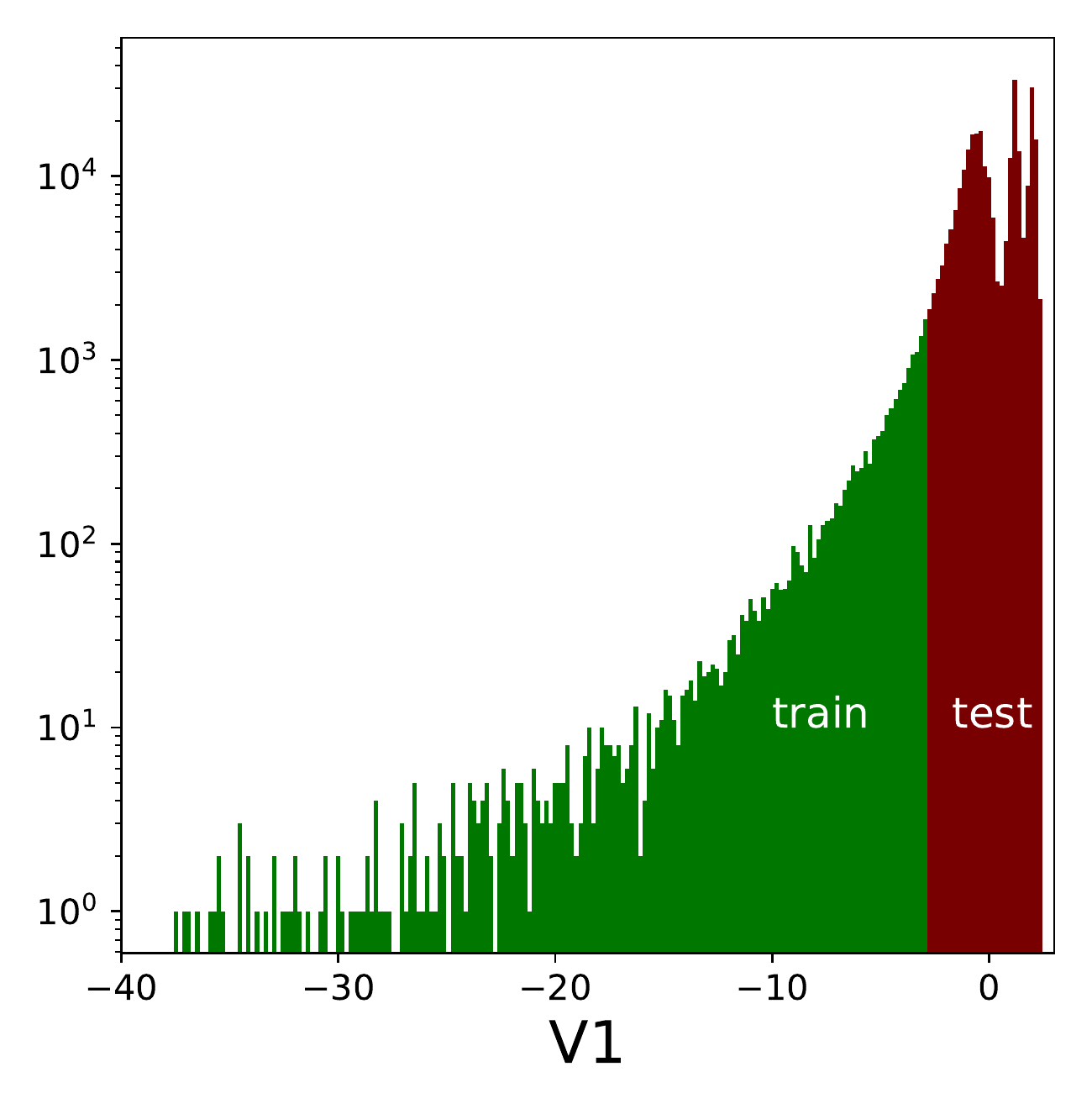}
		\newsubcap{Distribution of the main PCA component V1 from the Credit Card Fraud Detection dataset \cite{CreditCardFraudDetection}. We split this dataset in two parts, where one is used for training a neural network (in green) and the other one for testing on unusualness (in red) using the metrics from Section \ref{sec:method}.}
		\label{fig:creditcard_data}
	\end{subfigure}%
	~
	\begin{subfigure}[t]{0.45\textwidth}
		\centering
		\includegraphics[width=\textwidth]{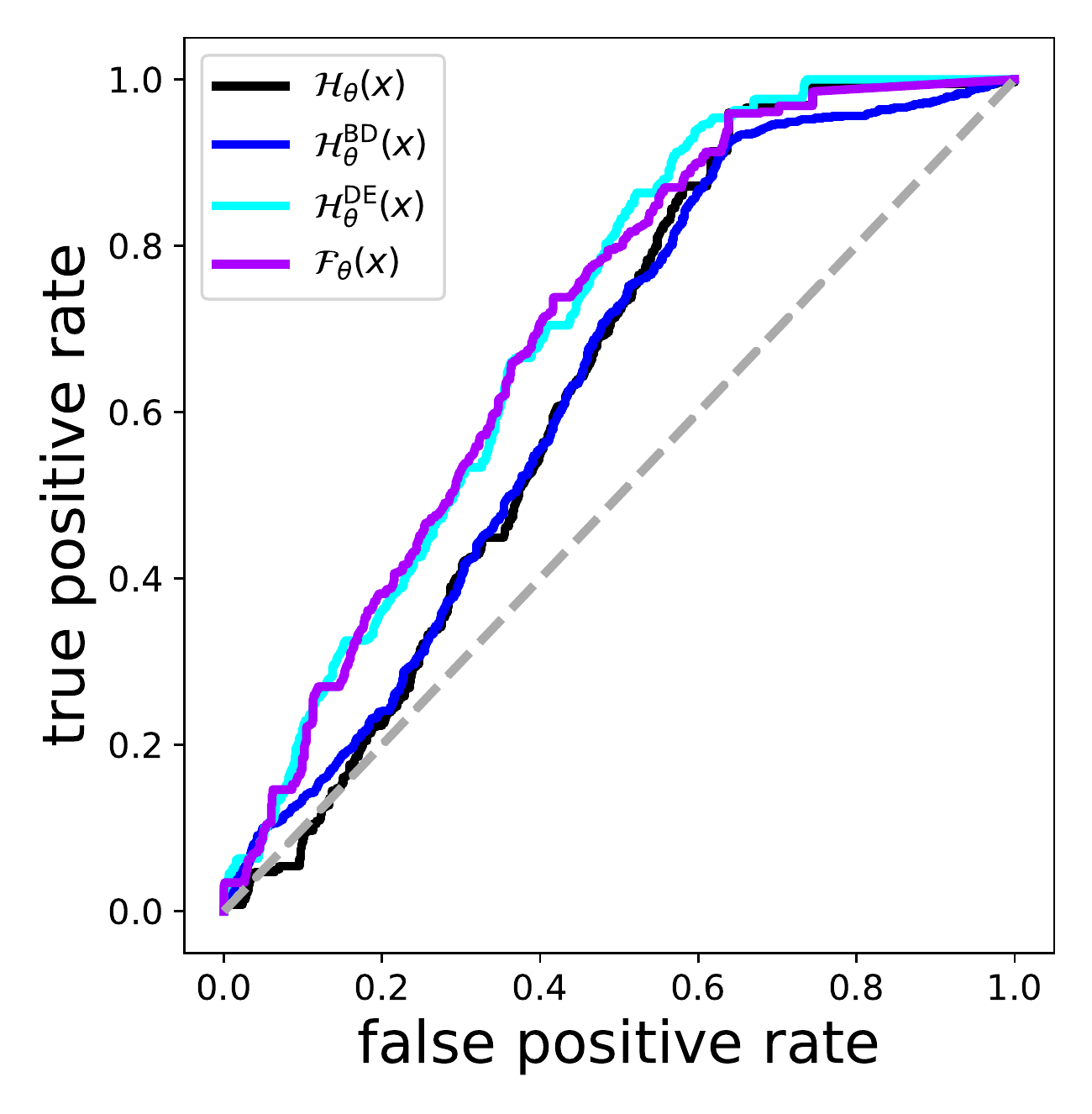}
		\newsubcap{ROC for detecting whether a datapoint is from the red or green area in Figure \ref{fig:creditcard_data} using the entropy (in black), the uncertainty computed from a deep ensemble (in blue) and the Fisher form (in purple).}
		\label{fig:creditcard_roc_1}
	\end{subfigure}
\end{figure*}
While in Section \ref{subsec:modified_data} we looked at the behavior of $\mathcal{F}_\theta$, $\mathcal{H}_\theta$, $\mathcal{H}_\theta^{\mathrm{DE}}$ and $\mathcal{H}_\theta^{\mathrm{GD/DE}}$ for modified data, we will now analyze another scenario. What happens if the data we used in training was \emph{incomplete}, in the sense that it lacked one or several important features that will occur in its application after training \cite{Amodei2016, Doshi2017, Snoek2019}?

As a first example we use the \emph{Credit Card Fraud Detection} dataset \cite{CreditCardFraudDetection, Dal2014}. This, extremely unbalanced, dataset contains 284,807 transactions with 492 frauds which are indicated by a binary label. The data were anonymized using a principal component analysis. The distribution of the main component, dubbed `V1`, is shown in Figure \ref{fig:creditcard_data}. We split the data in two halfs. A training set, where V1 is below a threshold of -3.0 (in green), and a test set where V1 is bigger than the threshold (in red). On the green part we trained, balancing the classes, a small fully connected neural network (Fig. \ref{fig:credit_net}) for 20 epochs until it reached an accuracy of $95\%$ on a smaller subset of the ``train'' set that was excluded from the training process. On the ``test'' set from Figure \ref{fig:creditcard_data} the accuracy is with $79\%$ markedly lower.  
\begin{figure*}[t]\centering
	\begin{subfigure}[t]{0.32\textwidth}
		\centering
		\includegraphics[width=\textwidth]{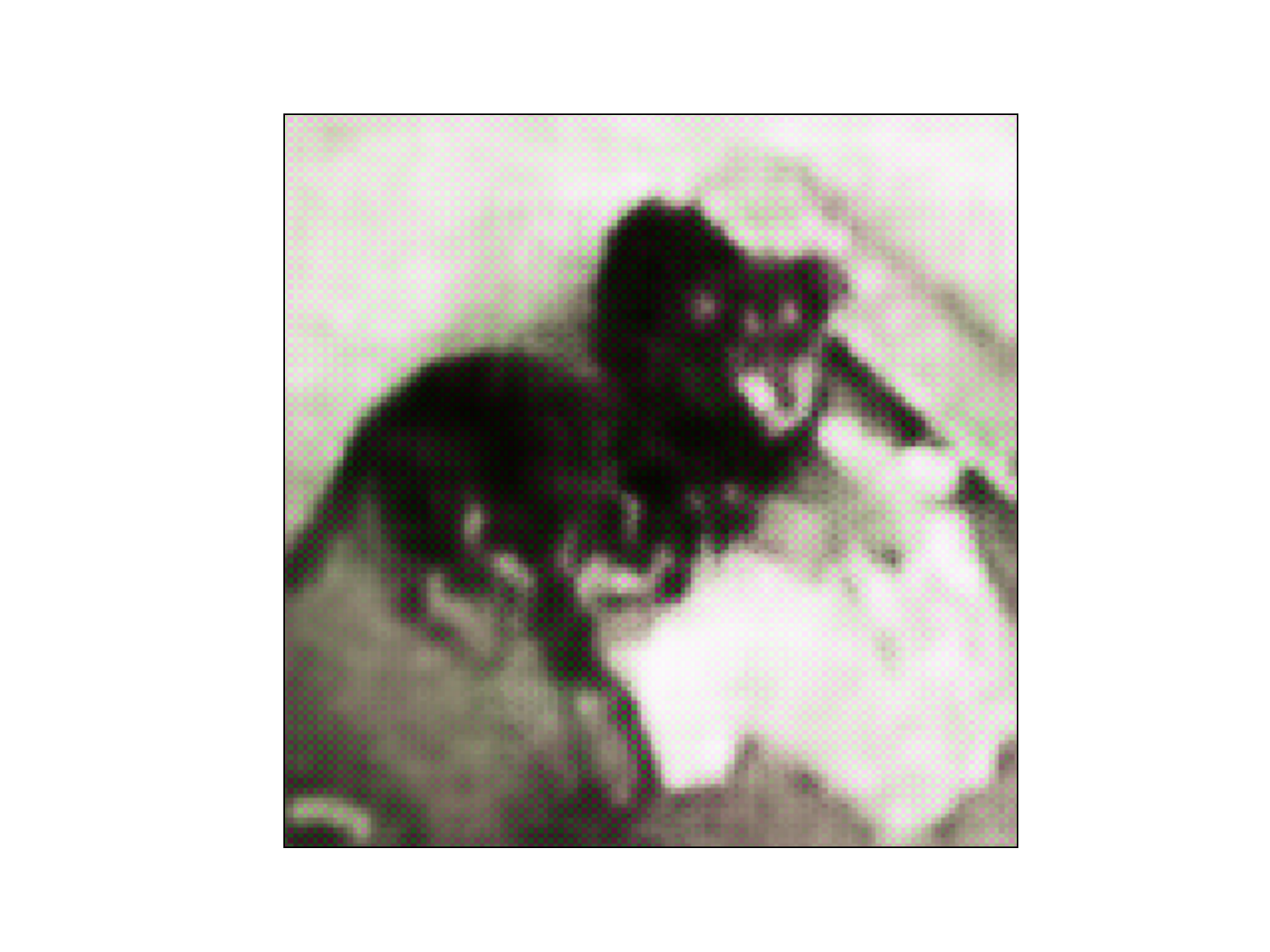}
		\newsubcap{Output of the used VAE for a dog image from DogsVsCats dataset \cite{DogsVsCats} where the variable in latent space was taken to be the mean of the distribution of the bottleneck node $\mathrm{En}$. For this image the bottleneck node $\mathrm{En}$ is larger than the threshold 0.3.}
		\label{fig:dog}
	\end{subfigure}%
	~
	\begin{subfigure}[t]{0.32\textwidth}
		\centering
		\includegraphics[width=\textwidth]{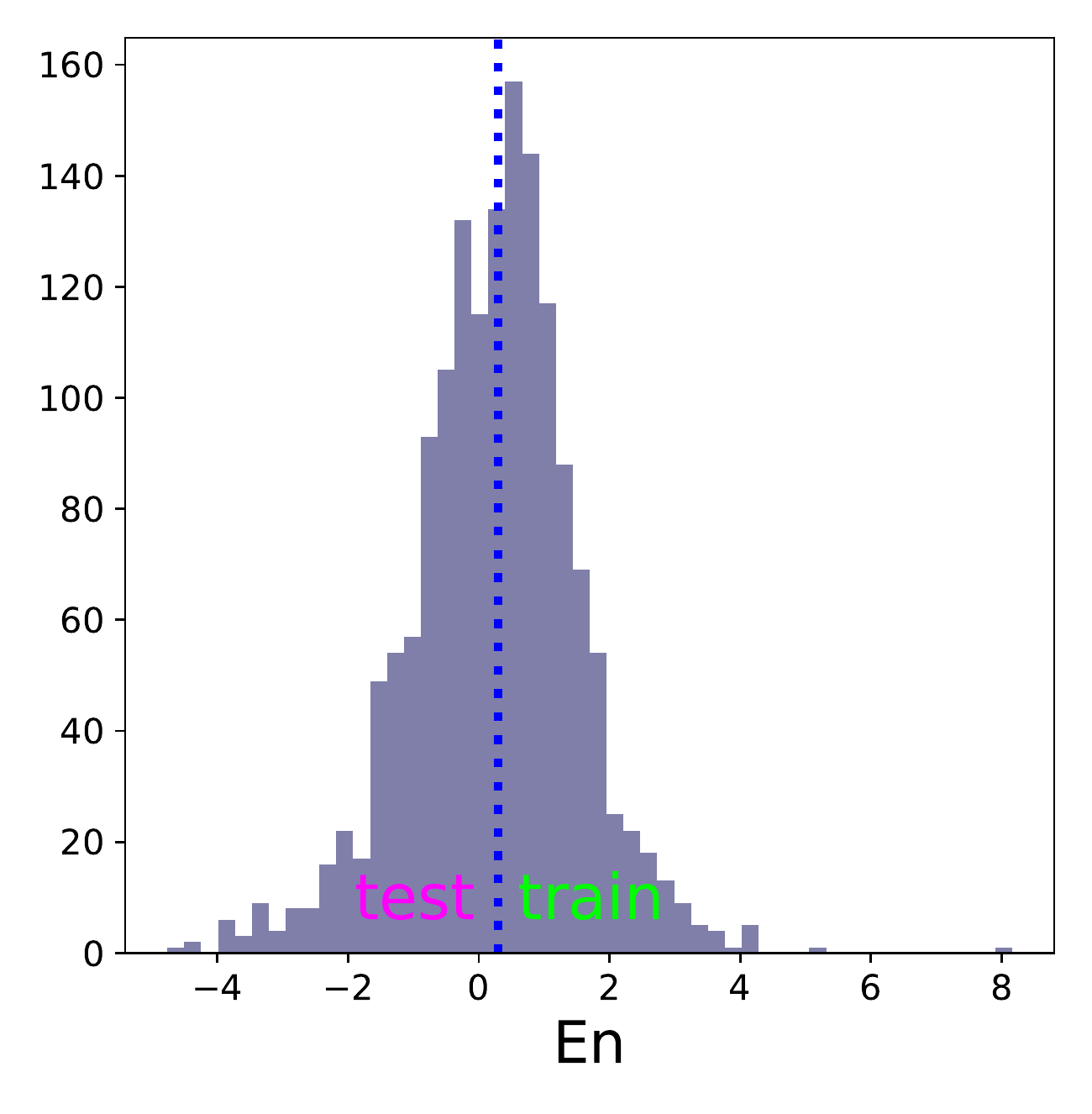}
		\newsubcap{Distribution of the bottleneck node $\mathrm{En}$ of the used VAE for dog images. $\mathrm{En}$ is the mean bottleneck node with largest variance within the dog images of the DogsVsCats dataset.}
		\label{fig:En}
	\end{subfigure}%
	~
	\begin{subfigure}[t]{0.32\textwidth}
		\centering 
		\includegraphics[width=\textwidth]{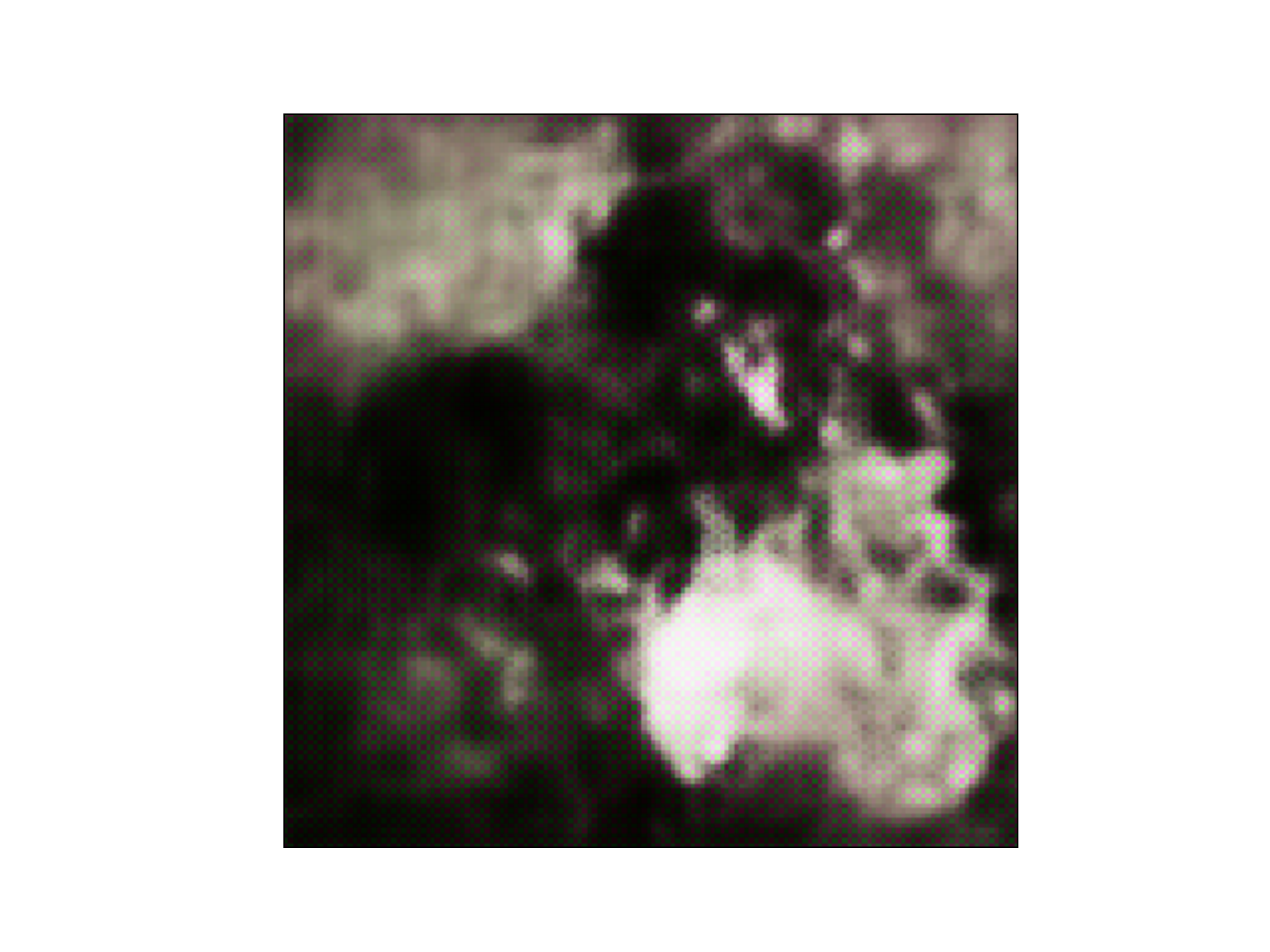} 
		\newsubcap{The output for the same as input as for Figure \ref{fig:dog} but with the sign of $\mathrm{En}$ flipped so that \eqref{eq:dog_unusual} is satisfied. Note that while the background in Fig. \ref{fig:dog} was bright, it is now turned dark.}
		\label{fig:changed_dog}
	\end{subfigure}
\end{figure*}
%

Figure \ref{fig:creditcard_roc_1} shows the ROC for the detection whether a datapoint $x$ is from the red part of Figure \ref{fig:creditcard_data} based on $\mathcal{H}_\theta(x)$, $\mathcal{F}_\theta(x)$, $\mathcal{H}_\theta^{\mathrm{DE}}(x)$ and $\mathcal{H}_\theta^{\mathrm{BD}}(x)$. The AUC for $\mathcal{F}_\theta$ and $\mathcal{H}_\theta^{\mathrm{DE}}$ is around 0.70, while $\mathcal{H}_\theta$ and $\mathcal{H}_\theta^{\mathrm{BD}}$ arrive both at $0.63$.

We will now consider a problem where the splitting is more subtle and, maybe, more realistic. For this purpose we will use the \emph{DogsVsCats} dataset from \cite{DogsVsCats}. We trained a variational autoencoder (Fig. \ref{fig:dog_vae}) \cite{Kingma2013} on the combined collection of dog images from \cite{DogsVsCats, CatandDog} for 30 epochs until it reached a mean squared error accuracy of around 2\% of the pixel range.
A reconstructed image can be seen in Figure \ref{fig:dog}. We then selected the bottleneck node, called $\mathrm{En}$ from now on, that exhibits the largest variance when evaluated on these dog images.
Figure \ref{fig:En} shows the distribution of $\mathrm{En}$ for dog images from \cite{DogsVsCats} together with $\mathrm{median}(\mathrm{En}) = 0.3$. 
What is the meaning of $\mathrm{En}$? Figure \ref{fig:changed_dog} shows the same reconstructed image as in \ref{fig:dog} but with flipped sign of $\mathrm{En}$.

We can see that the background of the depicted image becomes darker. 
Looking at various datapoints from the dataset hardens this impression so that we will see this as the ``meaning'' of $\mathrm{En}$, although there is, probably, no one-to-one correspondence with a human interpretation.
We trained a convolutional neural network (Fig. \ref{fig:dog_net}) on the data from \cite{DogsVsCats}, but by omitting those dog images with 
\begin{align}
	\label{eq:dog_unusual}
\mathrm{En} < \mathrm{median}(\mathrm{En}) = 0.3 \,,
\end{align}
i.e. those on the left side of the median in Figure \ref{fig:En} or, with the interpretation above, those images with dark background. We trained this network for 20 epochs until it reached an accuracy of around $80\%$ on a subset of the images with $\mathrm{En} \geq \mathrm{median}(\mathrm{En})$ that we ignored while training.
Evaluating now the network on dog images from \cite{DogsVsCats, CatandDog} with $\mathrm{En} < \mathrm{median}(\mathrm{En})$ lets the accuracy drop to 1\%! In other words almost all dogs images with $\mathrm{En}<0.3$ are classified as cats.
\begin{figure}[h]
	\includegraphics[width=0.48\textwidth]{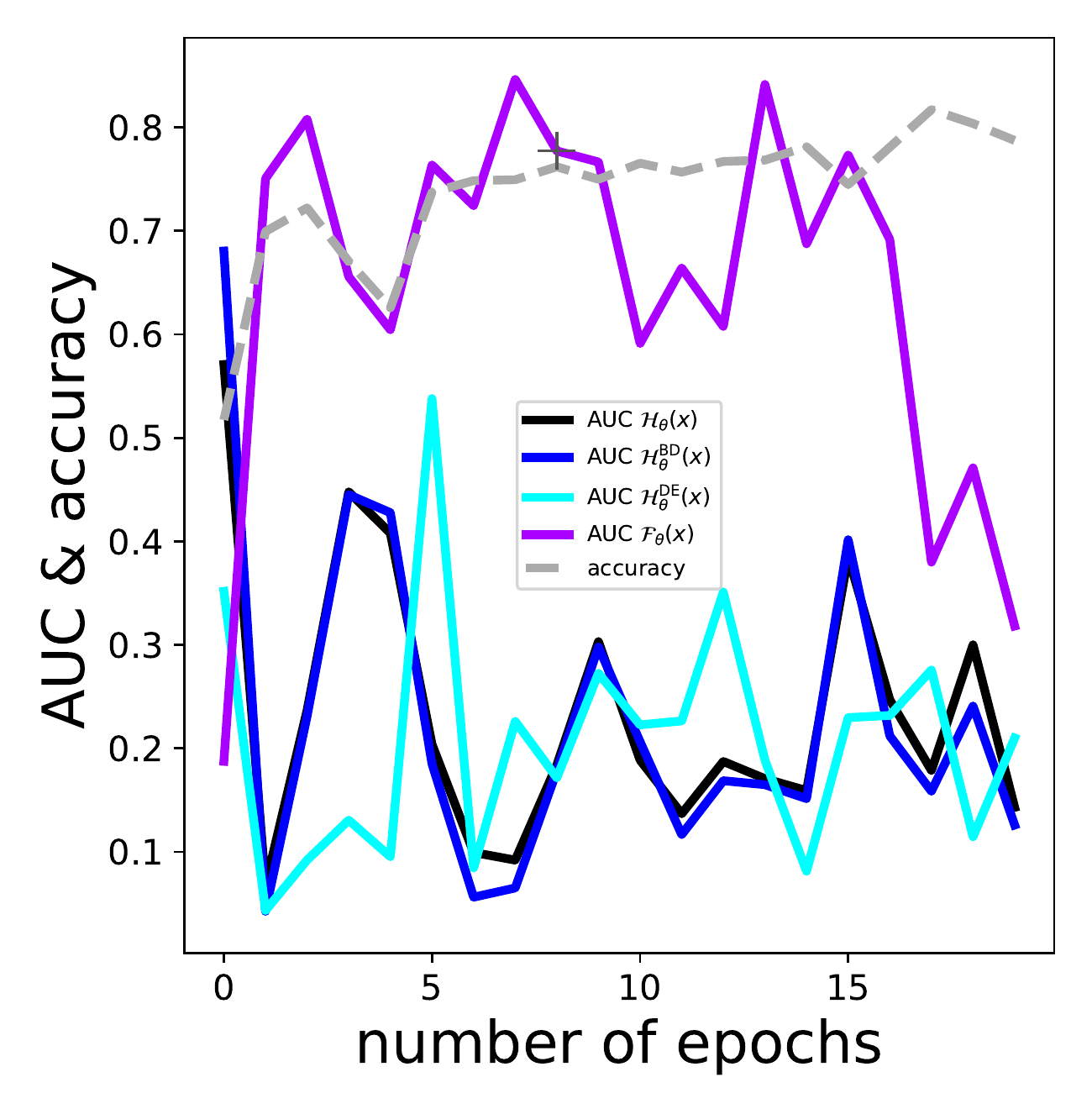}
	\newsubcap{Evolution the AUC for detecting whether a dog image from the DogsVsCats datset \cite{DogsVsCats} is ``unusual'' in the sense of \eqref{eq:dog_unusual} while training on datapoints where \eqref{eq:dog_unusual} is wrong. As metrics for detecting we use the entropies $\mathcal{H}_\theta(x)$, $\mathcal{H}_\theta^{\mathrm{BD}}(x)$, $\mathcal{H}_\theta^{\mathrm{DE}}(x)$ (in black, blue and cyan) and the Fisher form $\mathcal{F}_\theta(x)$ (in purple). The dashed gray line shows the accuracy of the trained network.}
	\label{fig:auc_evolution}
\end{figure}
Our task is now to detect whether an input is \emph{unusual} in the sense that \eqref{eq:dog_unusual} holds. 
Figure \ref{fig:auc_evolution} depicts the evolution of the AUC for the quantities $\mathcal{F}_\theta$, $\mathcal{H}_\theta$, $\mathcal{H}_\theta^{\mathrm{DE}}$ and $\mathcal{H}_\theta^{\mathrm{BD}}$ for each epoch and the accuracy of the trained network (in dashed gray). The difference between the Fisher form $\mathcal{F}_\theta$ and all other forms is striking. The quantities based on entropy seem to completely fail in detecting the ``unusualness'' of images with $\mathrm{En}<0.3$. In fact they yield AUCs below 0.5, thereby marking the actual training images as more suspicious than the modified images on which the classification actually fails. The Fisher form $\mathcal{F}_\theta$ however is yielding a value well above 0.5 for a wide range of the training. After around 15 epochs the AUC for the Fisher form starts to drop down as well.
\begin{figure*}[t]
	\begin{subfigure}[t]{0.45\textwidth}
		\centering
		\includegraphics[width=\textwidth]{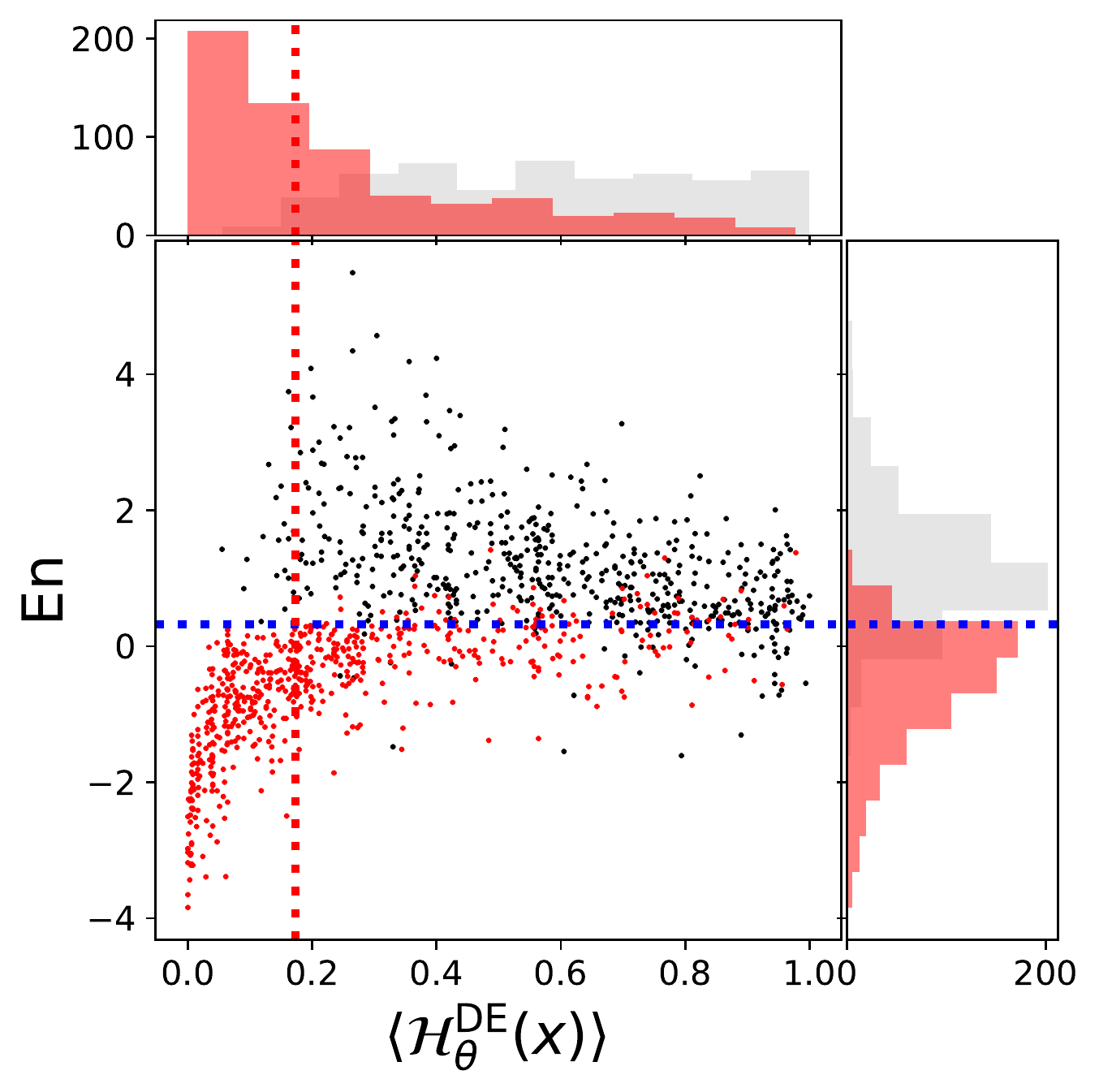}
	\end{subfigure}
	\begin{subfigure}[t]{0.45\textwidth}
		\centering
		\includegraphics[width=\textwidth]{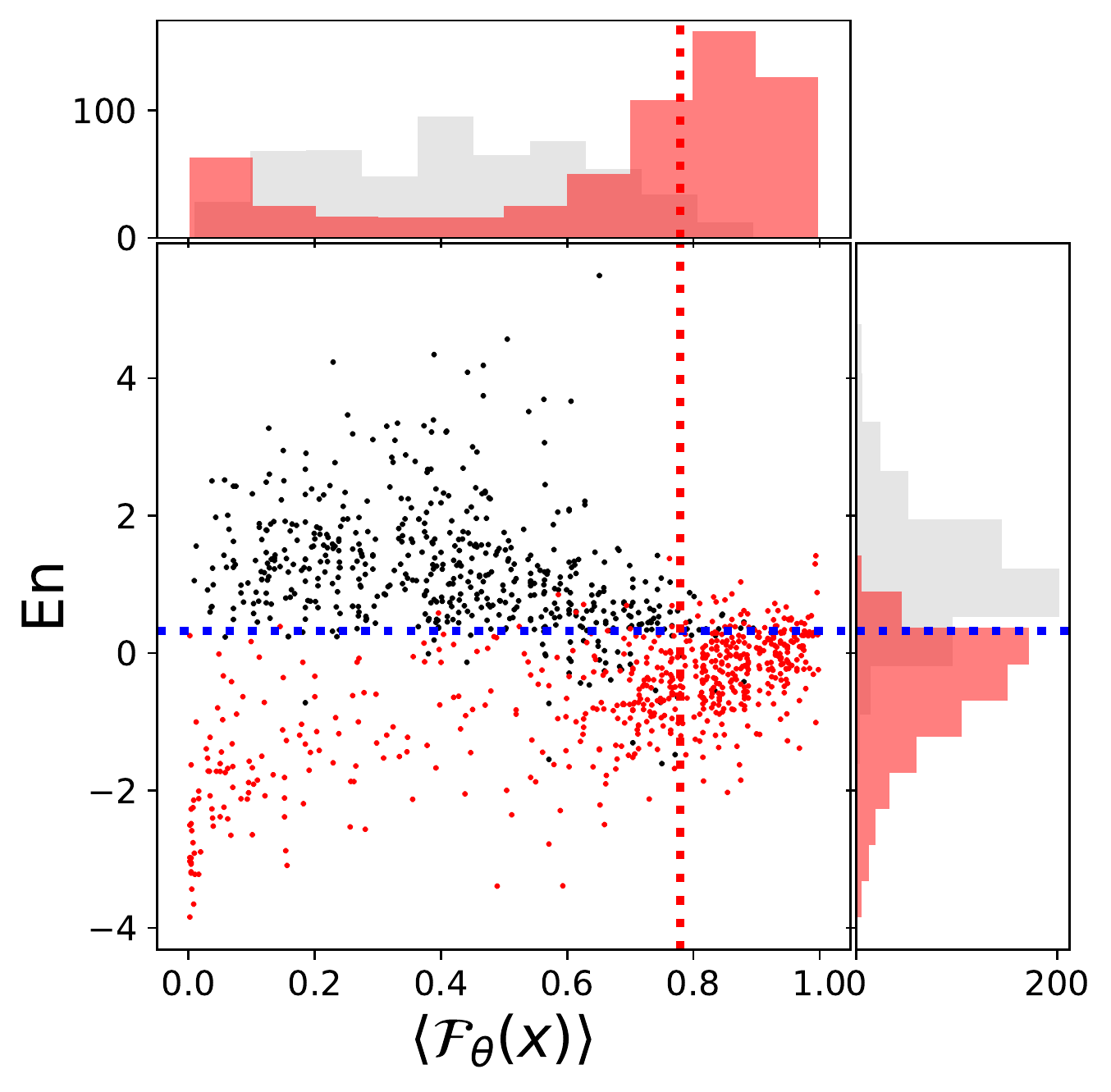}
	\end{subfigure}
	\newsubcap{The behavior of the entropy $\langle\mathcal{H}_\theta^{\mathrm{DE}}(x)\rangle$ predicted by the deep ensemble and the Fisher form $\langle\mathcal{F}_\theta(x)\rangle$ for dog images from \cite{DogsVsCats, CatandDog}. The ordinate shows the value of the bottleneck node $\mathrm{En}$ which was used to formulate the criterion \eqref{eq:dog_unusual}. Red marks those datapoints that were classified wrong (as a cat). The histograms show the distribution of the marginal, again split in correct and wrong classified datapoints. }
	\label{fig:dogs_cacc}
\end{figure*}
As this coincides roughly with the point where the accuracy starts stabilizing we suspect that this is a similar effect as we observed in Figure \ref{fig:intel_noise}: Once the maximal probability of the softmax output becomes almost identical to one, this forces the Fisher information to decrease. 
In fact it turns out that after 20 epochs the maximal probabilities on the unusual data are in average around $99\%$. Compare this to the tiny accuracy of $1\%$! We will focus now on a point before this, apparently unreasonable, saturation happened, namely on the trained state after 8 epochs, marked by a gray cross. 

The normalization from \eqref{eq:normalization}, for which we here used the training data, allows us to compare the behavior of $\mathcal{H}_\theta(x)$, $\mathcal{H}^{\mathrm{DE}}_\theta(x)$, $\mathcal{H}_\theta^{\mathrm{BD}}(x)$ and $\mathcal{F}_\theta(x)$ for single datapoints $x$. As all entropies show a quite a similar behavior for this example, we will only compare $\mathcal{H}_\theta^{\mathrm{DE}}$ and $\mathcal{F}_\theta$.
We drew 1200 images all showing dogs and depicted the value of $\mathrm{En}$ versus $\langle\mathcal{H}_\theta^{\mathrm{DE}}(x)\rangle$ and $\langle\mathcal{F}_\theta(x)\rangle$ in Figure \ref{fig:dogs_cacc}. Wrong classifications as cats are marked in red. The median $0.3$ of $\mathrm{En}$ is drawn as a blue dashed line, so that the ``unusual datapoints'' are below this line. The marginal distributions of the detection quantities and of $\mathrm{En}$ are shown, split according to classified correctly (in black) and wrongly (in red).

Note first, that indeed most of the images satisfying \eqref{eq:dog_unusual} are classified wrong, so that the network is in fact not suitable for classifying these datapoints. Moreover, the deep ensemble entropy is not able to capture that there is something off with these data. Even worse, the corresponding points are cumulated in the lower left corner which should actually indicate a high confidence in the prediction for these points. The Fisher information on the other hand leads to a cumulation in the lower right corner and thus correctly detects many of the ``unusual'' datapoints. In fact, even for ``normal'' datapoints we observe that most of those that were classified wrong have an encoder value $\mathrm{En}$ close to $0.3$ and possess again a high $\mathcal{F}_\theta$.

\section{Conclusion and outlook}
\balance
We studied the \emph{Fisher form} to detect whether an input is unusual, in the sense that it differs from the data that were used to infer the learned parameters. Spotting such a disparity is important, as it can substantially decrease the reliability. Several examples for this have been presented in Section \ref{sec:experiments}. We observed that the Fisher form performs equally or even better than competing methods in detecting unusual inputs. In particular, we introduced a normalization that allows to directly compare these metrics for single datapoints.

The last example treated in this article showed an effect, which could be a starting point for future research. As we observed in Figure \ref{fig:auc_evolution} the ability to detect unusual data might erode for a too long training. When only looking at the accuracy for an incomplete test set, as we did in Figure \ref{fig:auc_evolution}, such a process might happen unnoticed. This somehow different flavor of overfitting could be worth studying.

Another point that would deserve a more detailed investigation is of a more conceptual nature. 
For each of the examples treated in this work we used a single network to evaluate its Fisher form. One might wonder whether the effect we observed when considering unusual data will differ for another network whose training was carried out with, say, different initial conditions. As neural networks are bound to end up in local minima it is not obvious whether a datapoint will be unusual for two of these networks, even if they are trained on the same training set. However, this is not what we observed. In fact, the behavior of the Fisher form seems rather robust in this regard. It is hard to say, whether this indicates that neural networks tend to learn similar patterns when presented with the same data. Looking at the quadratic form \eqref{eq:FisherForm} for more than only one direction $v$ could help to shed a bit of light on this question.

\onecolumn
{
	\footnotesize
	\twocolumn
	\bibliography{unexpected}
	\bibliographystyle{ieeetr}
	\onecolumn
}

\appendix
\section*{Appendix}
\begin{figure}[h]\centering
	\begin{subfigure}[t]{0.38\textwidth}
		\centering
		\includegraphics[width=\textwidth]{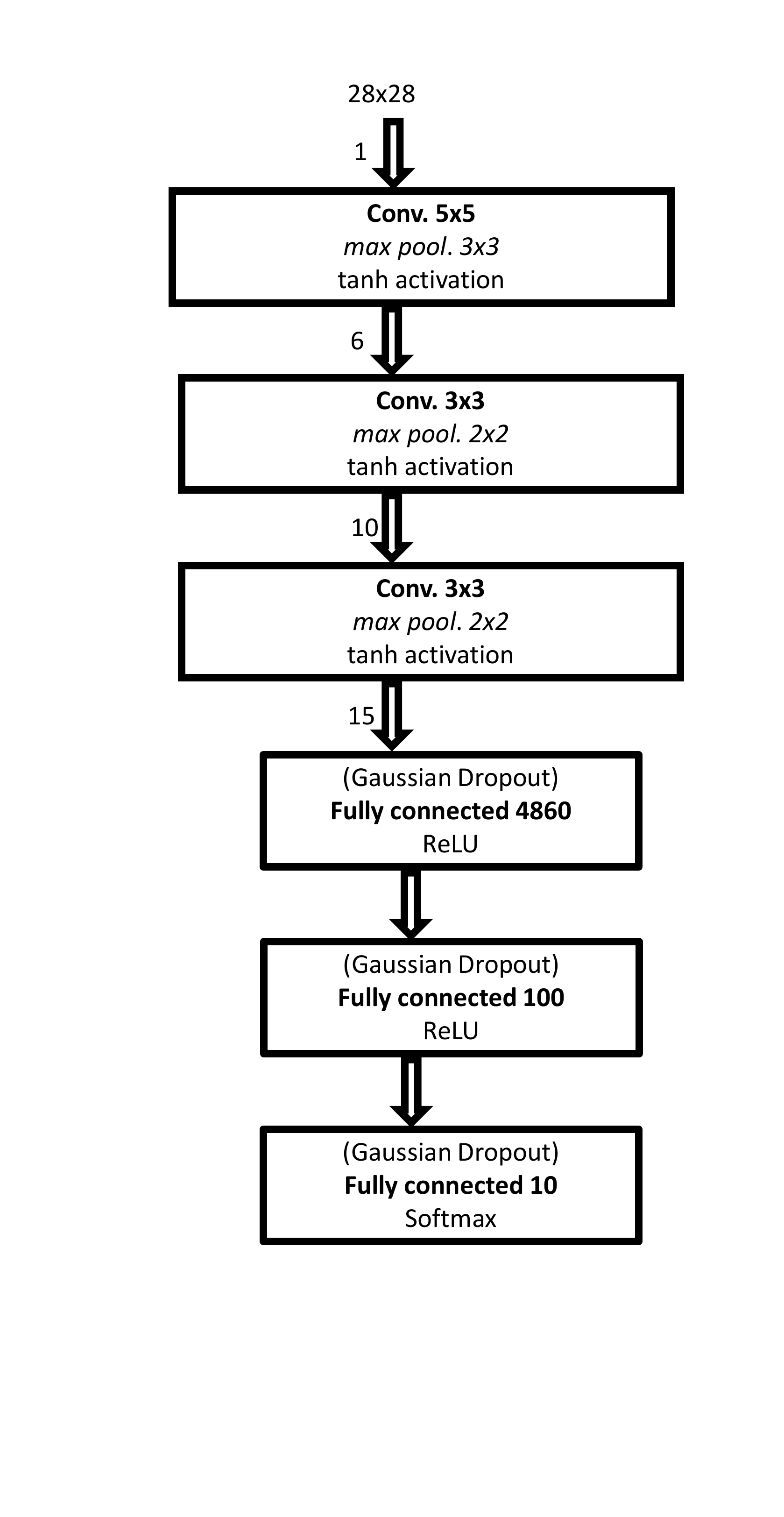}
		\newsubcap{Architecture for the convolutional neural network trained on the MNIST dataset that we use in Section \ref{subsec:modified_data}. For convolutional layers, we marked incoming channels by the number next to the arrow.}
		\label{fig:mnist_net}
	\end{subfigure}%
	~
	\begin{subfigure}[t]{0.38\textwidth}
		\centering
		\includegraphics[width=\textwidth]{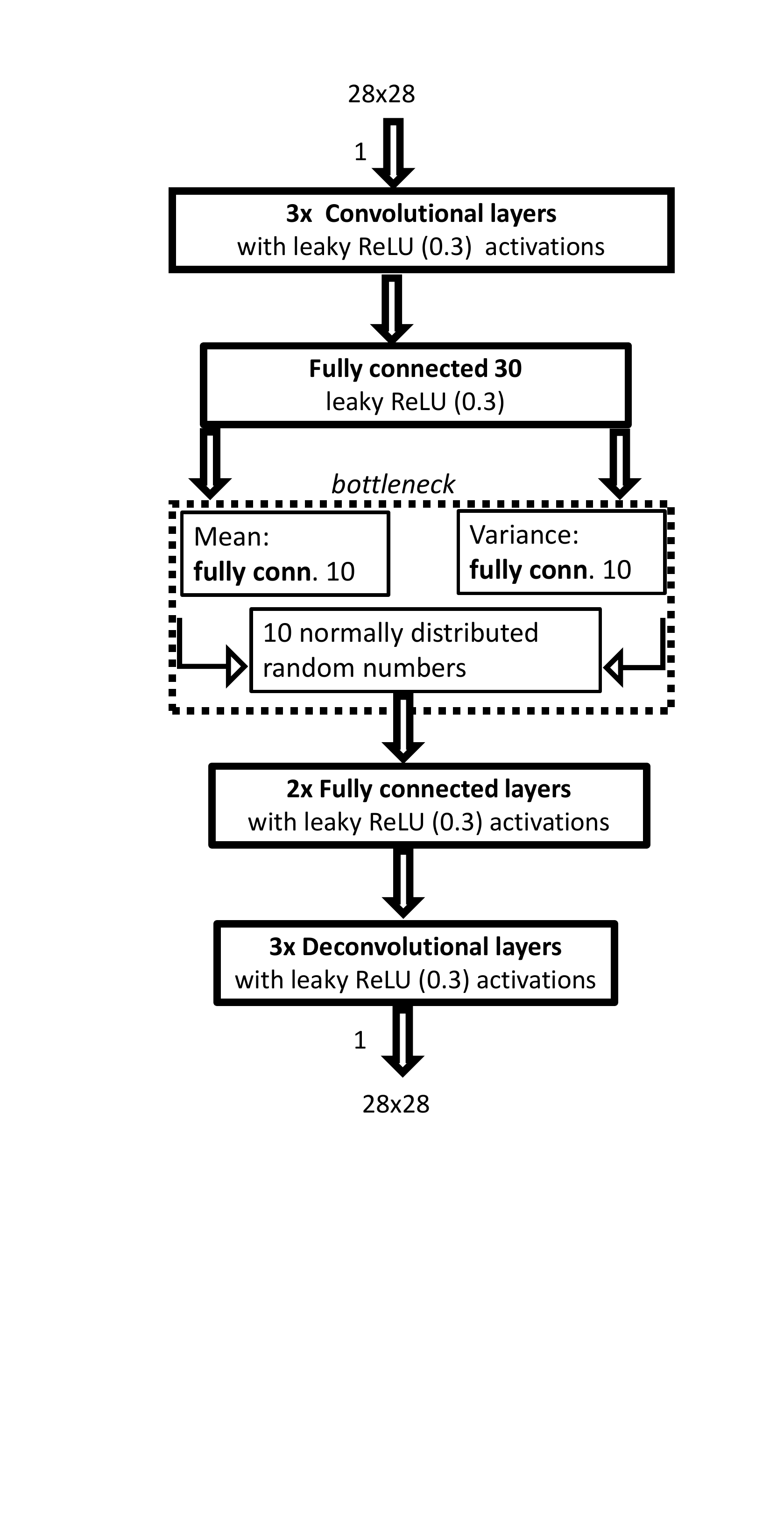}
		\newsubcap{Architecture for the variational autoencoder we train on the MNIST dataset and use in Section \ref{subsec:modified_data}. Note that the bottleneck is random: From the 10``means'' and 10 ``variances'' in the bottleneck we construct 10 Gaussian distributions from which we draw 10 random numbers that are than feeded through the decoder.}
		\label{fig:mnist_vae}
	\end{subfigure}
\end{figure}

\begin{figure}[t]\centering
	\begin{subfigure}[t]{0.38\textwidth}
		\centering
		\includegraphics[width=\textwidth]{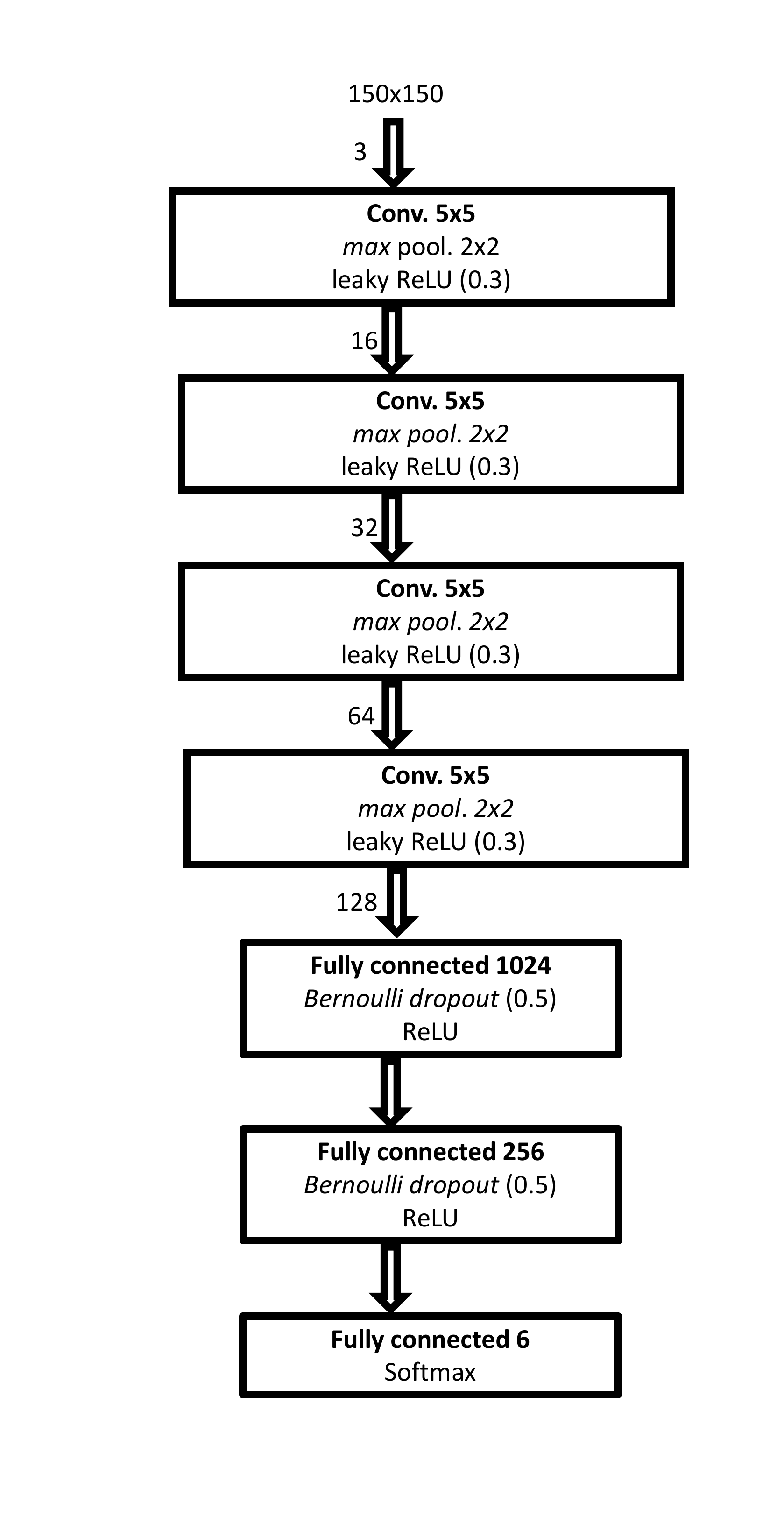}
		\newsubcap{Architecture for the convolutional neural network trained on the Intel Image classification dataset that we use in Section \ref{subsec:modified_data}. For convolutional layers, we marked incoming channels by the number next to the arrow.}
		\label{fig:intel_net}
	\end{subfigure}%
	~
	\begin{subfigure}[t]{0.38\textwidth}
		\centering
		\includegraphics[width=\textwidth]{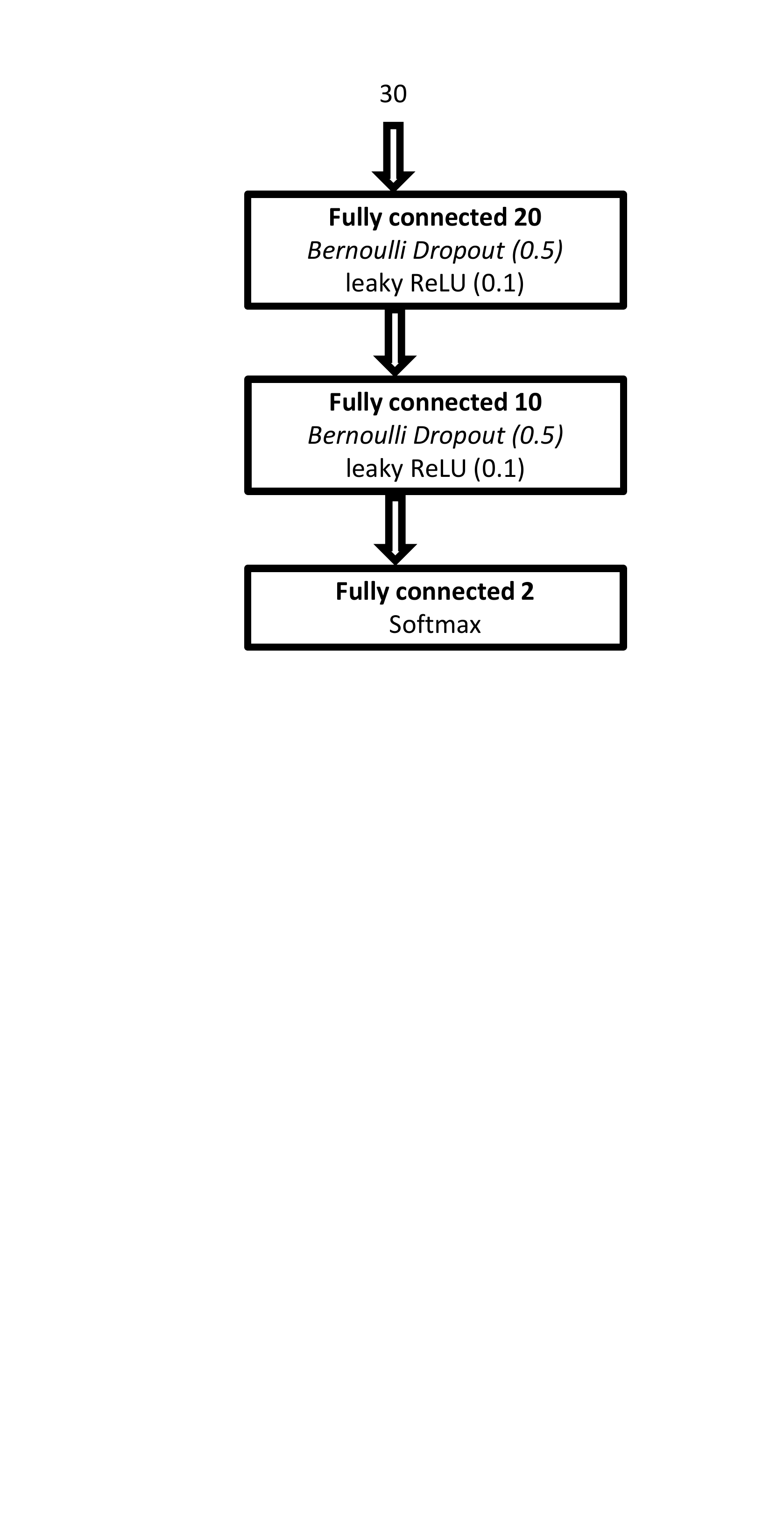}
		\newsubcap{Architecture for the fully connected neural network trained on the Credit Card Fraud Detection dataset that we use in Section \ref{subsec:incomplete_training_data}.}
		\label{fig:credit_net}
	\end{subfigure}
\end{figure}

\begin{figure}[t]\centering
	\begin{subfigure}[t]{0.38\textwidth}
		\centering
		\includegraphics[width=\textwidth]{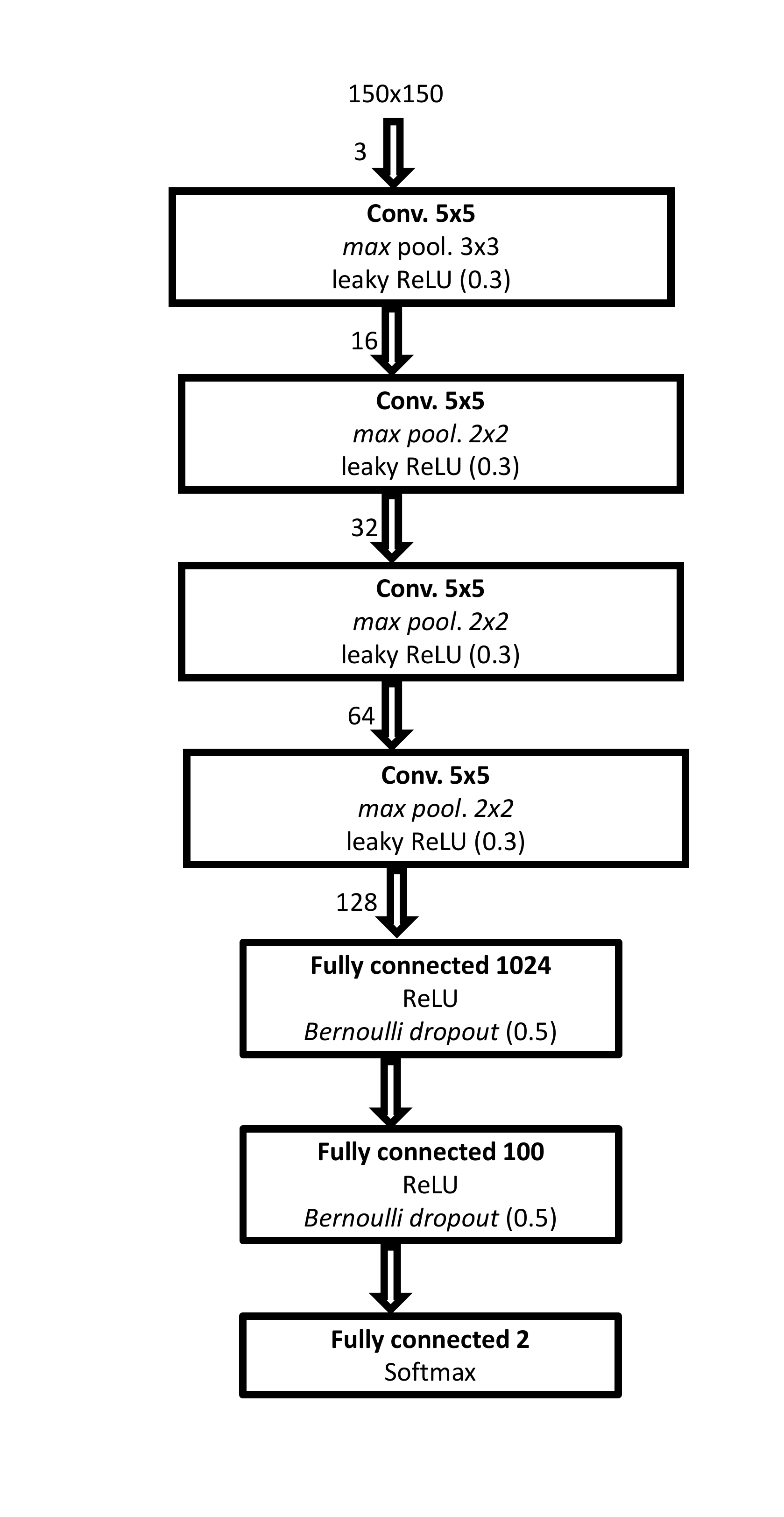}
		\newsubcap{Architecture for the convolutional neural network trained on the DogsVsCats dataset that we use in Section \ref{subsec:incomplete_training_data}. For convolutional layers, we marked incoming channels by the number next to the arrow.}
		\label{fig:dog_net}
	\end{subfigure}%
	~
	\begin{subfigure}[t]{0.38\textwidth}
		\centering
		\includegraphics[width=\textwidth]{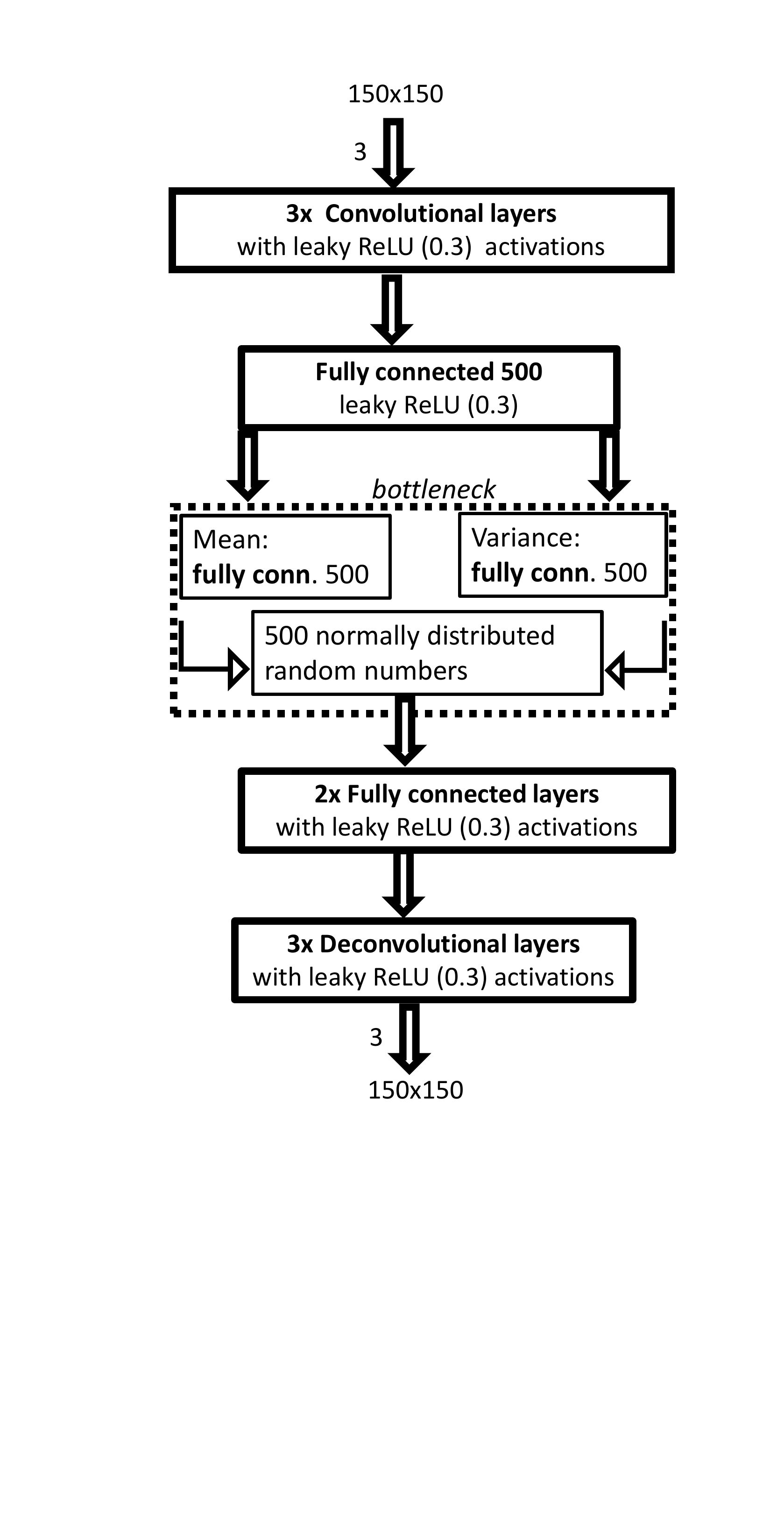}
		\newsubcap{Architecture for the variational autoencoder we train on the dog images from \cite{DogsVsCats, CatandDog} and use in Section \ref{subsec:incomplete_training_data}. Note that the bottleneck is random: From the 500``means'' and 500 ``variances'' in the bottleneck we construct 500 Gaussian distributions from which we draw 500 random numbers that are than feeded through the decoder.}
		\label{fig:dog_vae}
	\end{subfigure}
\end{figure}

\end{document}